\documentclass[letterpaper]{article} % DO NOT CHANGE THIS
\usepackage{aaai23}  % DO NOT CHANGE THIS
\usepackage{times}  % DO NOT CHANGE THIS
\usepackage{helvet}  % DO NOT CHANGE THIS
\usepackage{courier}  % DO NOT CHANGE THIS
\usepackage[hyphens]{url}  % DO NOT CHANGE THIS
\usepackage{graphicx} % DO NOT CHANGE THIS
\usepackage{natbib}  % DO NOT CHANGE THIS AND DO NOT ADD ANY OPTIONS TO IT
\usepackage{caption} % DO NOT CHANGE THIS AND DO NOT ADD ANY OPTIONS TO IT
\usepackage[noend]{algpseudocode}
\usepackage{algorithm}
\usepackage{algorithmicx}
\algrenewcommand\algorithmicindent{1.0em}
\usepackage{amsmath,amsthm,amssymb}
\usepackage{xspace}
\usepackage{multirow}
\usepackage{subfigure}

\newtheorem{example}{Example}
\newtheorem{theorem}{Theorem}
\newtheorem{lemma}[theorem]{Lemma}
\newtheorem{prop}[theorem]{Proposition}

\newtheorem{definition}{Definition}

\newcommand{\SOIRETM}{\texttt{SOIRETM}\xspace}
\newcommand{\SOIREDL}{\texttt{SOIREDL}\xspace}
\newcommand{\iSOIRE}{\texttt{iSOIRE}\xspace}
\newcommand{\GenICHARE}{\texttt{GenICHARE}\xspace}
\newcommand{\iSIRE}{\texttt{iSIRE}\xspace}
\newcommand{\RERNN}{\texttt{RE2RNN}\xspace}
\newcommand{\retree}{\texttt{RE2Tree}\xspace}
\newcommand{\treere}{\texttt{Tree2RE}\xspace}
\newcommand{\preorder}{\texttt{PreForm}\xspace}
\newcommand{\encre}{\texttt{Enc2Pre}\xspace}

\newcommand{\ie}{{\it i.e.}\xspace}

\newcommand{\resp}{{\it resp.}\xspace}
\newcommand{\etc}{{\it etc.}\xspace}

\title{A Noise-tolerant Differentiable Learning Approach for Single Occurrence Regular Expression with Interleaving}
\author{
    Rongzhen Ye, \textsuperscript{\rm 1}
    Tianqu Zhuang, \textsuperscript{\rm 1}
    Hai Wan, \textsuperscript{{\rm 1}*}
    Jianfeng Du, \textsuperscript{\rm 2}\thanks{Both Hai Wan and Jianfeng Du are corresponding authors.}
    Weilin Luo, \textsuperscript{\rm 1}
    Pingjia Liang, \textsuperscript{\rm 1}
}
\affiliations{
    \textsuperscript{\rm 1} School of Computer Science and Engineering, Sun Yat-sen University\\
    \textsuperscript{\rm 2} Guangzhou Key Laboratory of Multilingual Intelligent Processing, Guangdong University of Foreign Studies\\
    yerzh@mail2.sysu.edu.cn, zhangzhq58@mail2.sysu.edu.cn, wanhai@mail.sysu.edu.cn, jfdu@gdufs.edu.cn, luowlin3@mail2.sysu.edu.cn, liangpj3@mail2.sysu.edu.cn
}

\begin{document}

\maketitle

\begin{abstract}
We study the problem of learning a \emph{single occurrence regular expression with interleaving} (SOIRE) from a set of text strings possibly with \emph{noise}. 
SOIRE fully supports interleaving and covers a large portion of regular expressions used in practice.
Learning SOIREs is challenging because it requires heavy computation and text strings usually contain noise in practice.
Most of the previous studies only learn restricted SOIREs and are not robust on noisy data.
To tackle these issues, we propose a noise-tolerant differentiable learning approach \SOIREDL for SOIRE. 
We design a neural network to simulate SOIRE matching and theoretically prove that certain assignments of the set of parameters learnt by the neural network, called \emph{faithful encodings}, are one-to-one corresponding to SOIREs for a bounded size.
Based on this correspondence, we interpret the target SOIRE from an assignment of the set of parameters of the neural network by exploring the nearest faithful encodings.
Experimental results show that \SOIREDL outperforms the state-of-the-art approaches, especially on noisy data.
\end{abstract}

\section{Introduction}
\label{introduction}

Learning regular expressions (REs) is a fundamental task in Machine Learning. 
For example, REs are the target in 
EXtensible Markup Language (XML) schema inference, for covering a set of text strings about an XML element.
The regular expression with interleaving, denoted as RE(\&), is an extension of regular expressions, where the operator interleaving (\&) is added to interleave two strings.
RE(\&) has been widely used in various areas, ranging from XML database system~\cite{relaxng, Colazzo2013Efficient, Wim2017BonXai} to system verification~\cite{Gischer1981Shuffle, Mikolaj2006Two} and natural language processing~\cite{Kuhlmann2009Treebank, Joakim2009Projective}, \etc

We focus on a subclass of RE(\&), \emph{single occurrence regular expression with interleaving} (SOIRE), where each symbol occurs at most once in a SOIRE.
Learning SOIREs is still meaningful, as SOIREs have the second highest coverage rate of REs on the schema database Relax NG among all well-known subclasses~\cite{Yeting2019koire}. Although a subclass is generally easier to learn than the full class, it is still challenging to learn SOIREs.
On one hand, learning SOIREs is a search problem requiring heavy computation.
On the other hand, real-life text strings usually contain noise~\cite{Kearns1988Learning, Galassi2005Learning, Bex2006}.
For example, in XML schema inference the XML data may contain incorrect symbols~\cite{Bex2006}.
The presence of noise makes the problem of learning SOIREs more challenging.
%in DNA analysis, DNA sequences may be modified by insertion and deletion, leading to incorrect labels~\cite{Galassi2005Learning}. 

There have been a number of proposals for learning either the full class or its subclasses of SOIRE, from a set of text strings~\cite{Dominik2015Fast, Xiaolan2018Interleaving, Yeting2019effective, Yeting2020Negative}. 
However, they are hard to guarantee that the learnt REs reach the full declared expressive power.
For example, \citeauthor{Yeting2019effective}~\shortcite{Yeting2019effective} claimed to learn SOIREs, but \citeauthor{Xiaofan2021Discovering}~\shortcite{Xiaofan2021Discovering} showed that they just learn special cases of SOIRE.
Besides, existing approaches are not robust on noisy data because any modification to given strings will alter the patterns in learnt REs.
As far as we know, there is no approach to learning SOIREs that works well with both noise-free data and noisy data.

In this paper, we propose a noise-tolerant differentiable learning approach \SOIREDL for SOIREs.
Specifically, we design a neural network to simulate SOIRE matching for text strings. 
Since existing work for SOIRE matching~\cite{Xiaofan2021Shuffle, Xiaofan2022Membership} is not suitable for differentiable learning, we propose a new SOIRE matching algorithm \SOIRETM based on the syntax tree of SOIRE, and accordingly, an algorithm for converting \SOIRETM to a neural network.
We theoretically prove that certain assignments of the set of parameters learnt by the neural network, called \emph{faithful encodings}, one-to-one correspond to SOIREs for a bounded size.
This correspondence allows us to interpret the target SOIRE from an assignment of the set of learnt parameters of the neural network by exploring the nearest faithful encodings.

To evaluate the performance of \SOIREDL on noisy data, we extract $30$ SOIREs from the RE database built by \citeauthor{Yeting2018Practical}~\shortcite{Yeting2018Practical} to make a group of datasets covering five domains with different noise levels.
Experimental results show that at all noise levels, the average accuracy of \SOIREDL is higher than state-of-the-art (SOTA) approaches and the faithfulness of \SOIREDL is always beyond $80\%$. 
In particular, the average accuracy of \SOIREDL only decreases slightly with increasing noise levels, which suggests that \SOIREDL is robust on noisy data.

\section{Related Work}
\label{relatedwork}
  
{\bf Matching algorithms for regular expressions.}
It has been shown~\cite{Alain1994Complexity} that the matching problem of RE(\&) is NP-hard. 
For SOIREs, \citeauthor{Xiaofan2021Shuffle}~\shortcite{Xiaofan2021Shuffle} proposed a finite automata with interleaving, written FA(\&), to solve the matching problem.
%but FA(\&) can only be built from text strings instead of a given SOIRE. 
\citeauthor{Xiaofan2022Membership}~\shortcite{Xiaofan2022Membership} proposed a single occurrence finite automata, written SFA(\&, \#), for matching single occurrence regular expressions with interleaving and counting. All above studies do not consider converting the matching algorithm into a neural network. 
In contrast, we present not only a matching algorithm for SOIREs but also the way to convert it to a neural network.

\noindent {\bf Learning approaches for regular expressions.}
There also exists work for learning different subclasses of REs from a set of text strings, such as the deterministic regular expressions with counting~\cite{Xiaofan2018Counting}, the deterministic regular expressions with unorder~\cite{Xiaofan2020Unorder}, and the deterministic regular expression with counting and unorder~\cite{Xiaofan2021UnorderCounting}. 
Some subclasses of the extension RE(\&) have also been explored, such as chain regular expression with interleaving (ICHARE)~\cite{Xiaolan2018Interleaving}, restricted SOIRE (RSOIRE)~\cite{Yeting2019effective}, and k-occurrence regular expression with interleaving (kOIRE)~\cite{Yeting2020FlashSchema}. 
All of them are learnt from positive strings only.
\citeauthor{Yeting2020Negative}~\shortcite{Yeting2020Negative} learnt a subclass of ICHARE, called SIRE, from both positive and negative strings based on a genetic algorithm. 
The relation of expressive powers supported by these classes is SIRE $\subset$ ICHARE $\subset$ RSOIRE $\subset$ SOIRE $\subset$ kOIRE.
\citeauthor{Yeting2021TRANSREGEX}~\shortcite{Yeting2021TRANSREGEX} proposed a natural language processing based RE synthesizer to learn REs from natural language descriptions together with positive and negative strings. This problem setting is different from ours.

\noindent {\bf Differentiable learning.}
Differentiable learning has attracted much research interest recently.
Most studies focus on learning logical rules from knowledge bases~\cite{Fan2017Differentiable, Ali2019DRUM, William2020TensorLog, Jiani2021Scallop} or on neural logic programming~\cite{Yuan2020Learn, Kun2022Learning, Wang2020numerical}.
In particular, some studies~\cite{Tim2017Proving,Pasquale2020Learning,Pasquale2020Differentiable} focus on neural theorem proving. They convert the symbolic operations into differentiable modules to enhance the reasoning ability of the neural network. 
\citeauthor{Arthur2018Differentiable}~\shortcite{Arthur2018Differentiable} utilized a strongly convex regularizer to smooth the max operator and convert a broad class of dynamic programming (DP) algorithms into differentiable operators. 
\citeauthor{Wang2019SATNet}~\shortcite{Wang2019SATNet} proposed a differentiable MaxSAT solver integrated into the deep learning networks to solve the problems like visual Sudoku, which has implicit satisfiability constraints. 
For REs, \citeauthor{Chengyue2020Cold}~\shortcite{Chengyue2020Cold} injected a weighted finite-state automaton (FSA) of REs into the recurrent neural network (RNN) to improve the performance of text classification. 
Further, \citeauthor{Chengyue2021Neuralizing}~\shortcite{Chengyue2021Neuralizing} injected a finite-state transducer of REs into RNN for slot filling. 
Both of the above methods fine-tune initial regular expressions given from the expert knowledge to obtain better results.
Different from all above studies, we further study the one-to-one correspondence between parameters of a neural network and SOIREs.

\section{Preliminaries}
\label{preliminaries}

A regular expression with interleaving, written RE(\&), over an alphabet $\Sigma$ is defined recursively as follows~\cite{Alain1994Complexity}:
\begin{equation}
\nonumber
r := \epsilon \big| a \big| r_1^* \big| r_1 \cdot r_2 \big| r_1 \& r_2 \big| r_1 | r_2
\end{equation}
where $\epsilon$ is empty string, the symbol $a \in \Sigma$, and $r_1, r_2$ are RE(\&).
The operator $*$ denotes Kleene-Star, $\cdot$ denotes concatenation (it can be omitted if there is no ambiguity), $\&$ denotes interleaving, $|$ denotes disjunction. 
The operators $?$ and $+$ are commonly used for repetition. They are defined as $r^? := r \big| \epsilon$ and $r^+ := r \cdot r^*$, respectively. The operator $\&$ for two strings $s_1, s_2$ is defined as follows:
\begin{equation}
\nonumber
s_1 \& s_2 := \left\{\begin{array}{cc}
    s_2  & \text{if } s_1=\epsilon \\ 
    s_1  & \text{if } s_2=\epsilon \\ 
    a(s_1' \& s_2) | b(s_1 \& s_2') & otherwise\\ 
    \end{array}\right.
\end{equation}
where $s_1=as_1'$, $s_2=bs_2'$, $a, b \in \Sigma$.

\begin{definition}[Single occurrence regular expression with interleaving (SOIRE)~\cite{Yeting2019effective}]\label{def:SOIRE}
SOIRE is a RE(\&) where each symbol occurs at most once in the expression.
\end{definition}

\begin{example}\label{exam:SOIRE}
$(a\&b)c^*$ is a SOIRE. 
$(a\&b)a^*$ is a RE(\&) but not a SOIRE, because the symbol $a$ occurs twice. 
\end{example}

Each SOIRE can be expressed by its prefix notation where operators are written in front of operands rather than written in the middle as the infix notation. 
By $\preorder(r)$ we denote the prefix notation of a SOIRE $r$.
For the SOIRE $r=(a\&b)c^*$ given in Example~\ref{exam:SOIRE}, $\preorder(r)$ is $\cdot \& a b*c$.

\begin{figure}[htbp]
\centering
\includegraphics[width=0.65\linewidth]{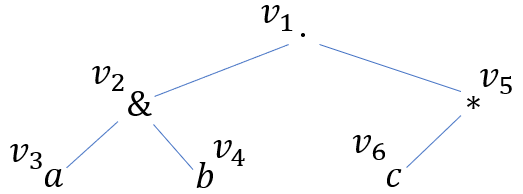}
\caption{The syntax tree of SOIRE $(a\&b)c^*$. $v_1$, \ldots, $v_6$ represent $\cdot$, $\&$, $a$, $b$, $*$ and $c$, respectively.}\label{fig:syntree}
\end{figure}

\noindent {\bf Syntax tree.}
Each SOIRE can also be represented as a binary tree, called \emph{syntax tree}, where each inner vertex in the tree represents an operator and each leaf represents a symbol. By $\retree(r)$ we denote the syntax tree of a SOIRE $r$. In a syntax tree, each leaf represents a symbol in $\Sigma$ and each symbol occurs at most once, while each inner vertex represents an operator and the number of children of it is equal to the number of its operands.
Figure~\ref{fig:syntree} shows the syntax tree of
$(a\&b)c^*$ in Example~\ref{exam:SOIRE}. 
The size of a SOIRE $r$, denoted by $|r|$, is defined as the number of vertices in the syntax tree of $r$. For example, the size of $(a\&b)c^*$ is $6$.
Obviously, the preorder traversal sequence of the syntax tree of $r$ is $\preorder(r)$ and each subtree represents a subexpression. 
In this preorder traversal sequence, vertex $t+1$ is the left child for any inner vertex $t$.
We use $r^t (1 \leq t \leq |r|)$ to denote the corresponding SOIRE of the subtree of $\retree(r)$ whose root is vertex $t$. Further, if vertex $t$ represents a binary operator, we use $\eta^t$ to denote the sequential number of its right child.
For $r=(a\&b)c^*$ in Example~\ref{exam:SOIRE}, $r^2=a\&b$ and $\eta^2=4$.

\noindent {\bf From prefix notation to syntax tree.}
% $\retree(r)$ builds the syntax tree from the bottom to top. 
We show a way to realize $\retree(r)$. 
It scans $\preorder(r)$ from back to front and
maintains a stack of syntax trees.
Take Figure~\ref{fig:syntree} for instance, $\preorder(r)=\cdot \& a b * c$.
$\retree(r)$ scans $\preorder(r)$ from $c$ to $\cdot$. 
When scanning $c$, push $c$ ($v_6$) into the stack.
When scanning $*$, pop $c$ ($v_6$) from the stack and make $c$ the left child of $*$, then push $*c$ ($v_5$) into the stack. 
When scanning $b$ and $a$, push $b$ ($v_4$) and $a$ ($v_3$) into the stack.
When scanning $\&$, pop $a$ ($v_3$) and $b$ ($v_4$) from the stack and make $a$ the left child of $\&$ and $b$ the right child, then push $\&ab$ ($v_2$) into the stack.
In this way, the syntax tree in Figure~\ref{fig:syntree} is built.  
We use $\treere(\xi)$ to denote the inverse function of $\retree(r)$, which returns $\preorder(r)$ from the syntax tree $\xi=\retree(r)$ by preorder traversal.

\noindent {\bf SOIRE matching.}
By $r \models s$ we denote that a SOIRE $r$ matches a string $s$. Then the problem of SOIRE matching is to check whether $r \models s$, which can be decided in the following way:
\begin{align}\label{equ:match}
    &r \models s := \nonumber\\
    &\left\{\begin{array}{ll}
    s=a  & \text{if } r=a\\ 
    s=\epsilon \vee r_1 \models s & \text{if } r=r_1^?\\ 
    s=\epsilon \vee \exists s_1 \exists s_2, (s=s_1 s_2 \wedge s_2 \neq \epsilon \\
    \wedge r_1^* \models s_1 \wedge r_1 \models s_2) & \text{if } r=r_1^*\\ 
    \exists s_1 \exists s_2, (s=s_1 s_2 \wedge r_1^* \models s_1 \wedge r_1 \models s_2) &  \text{if } r=r_1^+\\ 
    \exists s_1 \exists s_2, (s=s_1 s_2 \wedge r_1 \models s_1 \wedge r_2 \models s_2) & \text{if } r=r_1 \cdot r_2\\ 
    \exists s_1 \exists s_2, (s=s_1 \& s_2  \wedge r_1 \models s_1 \wedge r_2 \models s_2) & \text{if } r=r_1 \& r_2\\ 
    r_1  \models s \vee r_2 \models s & \text{if } r=r_1 | r_2 \\
    \end{array}\right.
\end{align}
where $r_1, r_2$ are SOIREs and $a \in \{\epsilon\} \cup \Sigma$.

\noindent {\bf SOIRE learning.}
Given a set of strings $\Pi=\Pi^+ \cup \Pi^-$, where $\Pi^+$ (\resp $\Pi^-$) denotes the set of strings with the positive (\resp negative) label, the problem of SOIRE learning is to find a SOIRE $r$ that maximizes $\mathrm{acc}(r)$ defined below.
\begin{eqnarray}\label{equ:acc}
\mathrm{acc}(r) = \frac{ |\{s | r \models s, s \in \Pi^+ \}| + |\{s | r \not\models s, s \in \Pi^- \}|}{|\Pi|}
\end{eqnarray}

\section{The Proposed \SOIREDL Approach}
\label{approach}

In this section, we first describe a new matching algorithm \SOIRETM, then show how to convert \SOIRETM to a neural network. 
Afterwards, we show correspondence between parameters of the neural network and SOIREs.
Finally, we show how to interpret the target SOIRE from the parameters.

First of all, we introduce a variant problem of 
SOIRE matching, called \emph{filter matching}.
Filter matching for a SOIRE $r$ and a string $s$ is to check if $r$ matches $filter(s, \alpha(r))$, where $\alpha(r)$ denotes the set of symbol in a SOIRE $r$, and the filter function $filter(s, V)$ returns a string that only retains symbols in $V$, where $V \subseteq \Sigma$.
For Example~\ref{exam:SOIRE} and $s=dbac$, 
the corresponding problem of filter matching is to check if $(a\&b)c^*$ matches $filter(dbac, \{a, b, c\})=bac$, as $\alpha((a\&b)c^*)=\{a, b, c\}$.

Given a SOIRE $r$ and a string $s$, we use $g_{i, j}^t \in \{0, 1\}$ ($1 \leq t \leq |r|$, $1 \leq i, j \leq |s|$) to denote whether $r^t$ matches $filter(s_{i, j}, \alpha(r^t))$, where $s_{i, j}$ denotes the substring of $s$ from $i$ to $j$, and where $s_{1, 0}=\epsilon$ specially.
If $r^t$ matches $filter(s_{i, j}, \alpha(r^t))$, then $g_{i, j}^t=1$, otherwise $g_{i, j}^t=0$. 
In particular, $g_{1, 0}^t$ denotes if $r^t$ matches $\epsilon$ since for all $t$ from $1$ to $|r|$, $filter(s_{1, 0}, \alpha(r^t))=\epsilon$.
For Example~\ref{exam:SOIRE} and $s=dbac$, $g_{1, 2}^2$ denotes if $a\&b$ matches $ba$ and $g_{1, 2}^2=1$.

\begin{table}[htbp]
\centering
\scalebox{0.8}{
\begin{tabular}{|l|l|}
    \hline
    SOIRE & Semantics of $r \models filter(s, \alpha(r))$\\
    \hline
    $r=a \in \Sigma$ & 1. $filter(s, {a})=a$. \\
    \hline
    $r=r_1^?$            & 1. $filter(s, \alpha(r_1^?))=\epsilon$. \\
    $=\epsilon | r_1$    & 2. $r_1 \models filter(s, \alpha(r_1))$. \\
    \hline
    $r=r_1^*$                             & 1. $filter(s, \alpha(r_1^*))=\epsilon$. \\
    $=\epsilon | r_1 | r_1^* \cdot r_1$   & 2. $r_1 \models filter(s, \alpha(r_1))$. \\
                                        & 3. $r_1^* \models filter(s_1, \alpha(r_1^*))$ and $r_1 \models filter(s_2, \alpha(r_1))$,\\
                                        &  where $s=s_1 s_2 (s_1, s_2 \neq \epsilon)$. \\
    \hline
    $r=r_1^+$                & 1. $r_1 \models filter(s, \alpha(r_1))$.\\
    $=r_1 | r_1^+ \cdot r_1$ & 2. $r_1^+ \models filter(s_1, \alpha(r_1^+))$ and $r_1 \models filter(s_2, \alpha(r_1))$, \\
                            &  where $s=s_1 s_2 (s_1, s_2 \neq \epsilon)$. \\
    \hline
    $r=r_1 \cdot r_2$        & 1. $r_1 \models filter(s, \alpha(r_1 \cdot r_2))$ and $r_2 \models \epsilon$. \\
                            & 2. $r_2 \models filter(s, \alpha(r_1 \cdot r_2))$ and $r_1 \models \epsilon$. \\
                            & 3. $r_1 \models filter(s_1, \alpha(r_1 \cdot r_2))$ and \\
                            & $r_2 \models filter(s_2, \alpha(r_1 \cdot r_2))$, where $s=s_1 s_2 (s_1, s_2 \neq \epsilon)$. \\
    \hline
    $r=r_1 \& r_2$ & $r_1 \models filter(s, \alpha(r_1))$ and $r_2 \models filter(s, \alpha(r_2))$.\\
    \hline
    $r=r_1 | r_2$ & 1. $r_1 \models filter(s, \alpha(r_1 | r_2))$. \\
                & 2. $r_2 \models filter(s, \alpha(r_1 | r_2))$. \\
    \hline
\end{tabular}}
\caption{The semantics of filter matching, where $r, r_1, r_2$ are SOIREs and $s, s_1, s_2$ are strings. $r \models filter(s, \alpha(r))$ if and only if at least one condition on the right is satisfied.}
\label{tab:semantics}
\end{table}

\subsection{SOIRE Matching by \SOIRETM}

We propose a new matching algorithm for SOIRE, named \SOIRETM, based on dynamic programming. 
Generally, \SOIRETM divides the original matching problem into smaller ones to conquer. 
%Due to the operator $\&$ in SOIRE, we need to know if a subexpression matches a subsequence of the given string. 
We observe that SOIRE matching can be simplified to filter matching, as shown in the following theorem.

\begin{theorem}\label{thm:match_cond}
Given a SOIRE $r$ and a string $s$, 
$r \models s$ iff $filter(s, \alpha(r))=s$ and $r \models filter(s, \alpha(r))$.\footnote{All the proofs of theorems/lemmas/propositions are provided in appendix B.} 
\end{theorem}

Theorem~\ref{thm:match_cond} presents a necessary and sufficient condition for SOIRE matching. 
The condition $filter(s, \alpha(r))=s$ guarantees that $\alpha(r)$ contains all symbols in $s$, which is easy to check.
Therefore, the primary problem is to check if $r$ matches $filter(s, \alpha(r))$. 

We build the syntax tree of $r$ and 
compute the result of filter matching of $s$ and $r$, namely $g_{1, |s|}^1$.
The proposed algorithm \SOIRETM is detailed in Algorithm~\ref{alg:SOIRETM}.
Initially, line~\ref{alg:SOIRETM_alpha} checks if $filter(s, \alpha(r))=s$.
Then we calculate $g_{i, j}^t$ from shorter substrings to longer ones and from bottom to top of the syntax tree.
The statements in Line~\ref{alg:SOIRETM_cond1}-\ref{alg:SOIRETM_opr7} conform to the semantics of each operator, given by Table~\ref{tab:semantics} and Lemma~\ref{lem:lr_match}. 

\begin{lemma}\label{lem:lr_match}
Given a SOIRE $r$ and a string $s$. 
If $r=r_1 \cdot r_2$ or $r=r_1 \& r_2$ or $r=r_1 | r_2$, then for all $i \in \{1, 2\}$, $r_i \models filter(s, \alpha(r))$ iff $filter(s, \alpha(r_i))=filter(s, \alpha(r))$ and $r_i \models filter(s, \alpha(r_i))$. 
\end{lemma}

The time complexity of Algorithm~\ref{alg:SOIRETM} is $O(|s|^3|r|)$. It is sound and complete according to the following theorem.

\begin{theorem}\label{thm:match}
Given a SOIRE $r$ and a string $s$, $r \models s$ iff $\SOIRETM(r, s)=1$. 
\end{theorem}

\begin{algorithm}[ht]
\caption{Matching Algorithm for SOIRE, \SOIRETM.}\label{alg:SOIRETM}
\textbf{Input}: A SOIRE $r$ and a string $s$.\\
\textbf{Output}: The answer of whether $r$ matches $s$.

\begin{algorithmic}[1]    
\If {$filter(s, \alpha(r)) \neq s$\label{alg:SOIRETM_alpha}}
    \State \textbf{Return} $0$;
\EndIf

\State Build the syntax tree of $r$ by $\retree(r)$;
\State Let $flag_{i, j}^{t, t'}$ denote $1[filter(s_{i, j}, \alpha(r^t))=filter(s_{i, j}, \alpha(r^{t'}))]$, where $1[\mu]=1$ iff $\mu$ is true;

\ForAll{substring $s_{i, j}$ from $\epsilon$ to $s$ (shorter to longer)}
    \For{$t \gets |r|$ downto $1$}
    \If{$r^t=a \in \alpha(r)$\label{alg:SOIRETM_cond1}}
        \State $g_{i, j}^t \gets 1[filter(s_{i, j}, \{a\})=a]$; \label{alg:SOIRETM_opr1}
    \ElsIf{$r^t=(r^{t+1})^?$}
        \State $g_{i, j}^t \gets 1[filter(s_{i, j}, \alpha(r^t))=\epsilon] \vee g_{i, j}^{t+1}$; \label{alg:SOIRETM_opr2}
    \ElsIf{$r^t=(r^{t+1})^*$}
        \State $g_{i, j}^t \gets 1[filter(s_{i, j}, \alpha(r^t))=\epsilon] \vee g_{i, j}^{t+1} \vee \bigvee_{k=i}^{j-1} (g_{i, k}^t \wedge g_{k+1, j}^{t+1})$; \label{alg:SOIRETM_opr3}
    \ElsIf{$r^t=(r^{t+1})^+$}
        \State $g_{i, j}^t \gets g_{i, j}^{t+1} \vee \bigvee_{k=i}^{j-1} (g_{i, k}^t \wedge g_{k+1, j}^{t+1})$; \label{alg:SOIRETM_opr4}
    \ElsIf{$r^t=r^{t+1} \cdot r^{\eta^t}$}
        \State $g_{i, j}^t \gets (flag_{i, j}^{t, t+1} \wedge g_{i, j}^{t+1} \wedge g_{1, 0}^{\eta^t}) \vee (flag_{i, j}^{t, \eta^t} \wedge g_{i, j}^{\eta^t} \wedge g_{1, 0}^{t+1}) \vee \bigvee_{k=i}^{j-1} (flag_{i, k}^{t, t+1} \wedge g_{i, k}^{t+1} \wedge flag_{k+1, j}^{t, \eta^t} \wedge g_{k+1, j}^{\eta^t})$; \label{alg:SOIRETM_opr5}
    \ElsIf{$r^t=r^{t+1} \& r^{\eta^t}$}
        \State $g_{i, j}^t \gets g_{i, j}^{t+1} \wedge g_{i, j}^{\eta^t}$; \label{alg:SOIRETM_opr6}
    \ElsIf{$r^t=r^{t+1} | r^{\eta^t}$\label{alg:SOIRETM_cond7}}
        \State $g_{i, j}^t \gets (flag_{i, j}^{t, t+1} \wedge g_{i, j}^{t+1}) \vee (flag_{i, j}^{t, \eta^t} \wedge g_{i, j}^{\eta^t})$; \label{alg:SOIRETM_opr7}
    \EndIf
    \EndFor
\EndFor

\State \textbf{Return} $g_{1, |s|}^1$;\label{alg:SOIRETM_return}

\end{algorithmic}
\end{algorithm}

\subsection{From \SOIRETM to Neural Network}

We now detail how to convert \SOIRETM to a trainable neural network to simulate SOIRE matching.

\SOIRETM uses the syntax tree for SOIRE matching, so the trainable parameters of an expected neural network can be defined by constructs of the syntax tree.
There are two parts of parameters used in an expected neural network, $\theta=(w, u)$, where $w \in [0, 1]^{T \times |\mathbb{B}|}$, $u \in [0, 1]^{T \times T}$ and $\mathbb{B}=\Sigma \cup \{?, *, +, \cdot, \&, |, \text{none}\}$, and where $T$ is the bounded size of the target SOIRE.
For $1 \leq t \leq T, a \in \mathbb{B}$, $w^t_a$ denotes the probability of vertex $t$ representing a symbol in $\Sigma$ or an ordinary operator or the none operator. For $1 \leq t \leq T$ and $t+2 \leq t' \leq T$,
$u^t_{t'}$ denotes the probability of vertex $t$ choosing vertex $t'$ as its right child. The total number of the parameters to be learnt is $T|\mathbb{B}|+\frac{(T-1)(T-2)}{2}$.
Example~\ref{exam:encoding} shows the parameters of the neural network $\theta=(w, u)$ corresponding to the syntax tree in Figure~\ref{fig:syntree}.

\begin{example}\label{exam:encoding}
When $T=6$, $w^1_{\cdot}$, $w^2_{\&}$, $w^3_{a}$, $w^4_{b}$, $w^5_{*}$, $w^6_{c}$, $u^1_{5}$, $u^2_{4}$ are $1$s, whereas other parameters are $0$s. 
When $T=8$, $w^{7}_{\text{none}}$, $w^{8}_{\text{none}}$ are $1$s in addition to the above parameters. 
\end{example}

There are four parts that should be considered during the conversion of \SOIRETM to neural network: $\alpha(r^t)$, $flag_{i, j}^{t, t'}, g^t_{i, j}$, and the return value of Algorithm~\ref{alg:SOIRETM}.

Recall that $\alpha(r^t)$ represents the set of symbols in $r^t$, which is also the set of symbols occurring in the subtree whose root is $t$. 
We use $\rho^t_a$ to denote the probability of symbol $a\in \Sigma$ that occurs in the subtree whose root is $t$.
We calculate $\rho^t_a$ from bottom to top of the syntax tree by Equation~\ref{equ:alpha}, where $\sigma_{01}(x)=\min(\max(x, 0), 1)$.
For all $t>T$ and $a\in \Sigma$, $\rho^t_a$ is set to $0$.
Note that $\min(x)=-\max(-x)$, and the $\max$ function amounts to ReLU and can be approximated by a more differentiable LeakyReLU function.
\begin{equation}\label{equ:alpha}
\rho^t_a=\sigma_{01}(w^t_a + \sum_{\substack{o \in \{?, *, +, \\ \cdot, \&, |\}}} w^t_o \rho^{t+1}_a + \sum_{\substack{o \in \\ \{\cdot, \&, |\}}} w^t_o \sum_{t'=t+2}^{T} u^t_{t'} \rho^{t'}_a)
\end{equation}

For converting $flag_{i, j}^{t, t'}$, we treat it as the probability that there does not exist a symbol occurring in both $s_{i, j}$ and $\alpha(r^t)$ but not occurring in $\alpha(r^{t'})$, as defined in Equation~\ref{equ:flag_t}. Note that $t'$ can only be either $t+1$ or $\eta^t$.
\begin{equation}\label{equ:flag_t}
flag_{i, j}^{t, t'}=1-\sigma_{01}(\sum_{a \in \Sigma} \sigma_{01}(1[a \in s_{i, j}]+(\rho^t_a-\rho^{t'}_a)-1))
\end{equation}

For converting $g_{i, j}^t$, we introduce 
$p_{i, j}^{t}(?)$ (\resp~$p_{i, j}^{t}(*)$ or $p_{i, j}^{t}(+)$) to denote the probability that the right-hand side of Line~\ref{alg:SOIRETM_opr2} (\resp~\ref{alg:SOIRETM_opr3} or \ref{alg:SOIRETM_opr4}) in Algorithm~\ref{alg:SOIRETM} evaluates to $1$, as well as $p_{i, j}^{t}(\cdot,t')$ (\resp~$p_{i, j}^{t}(\&,t')$ or $p_{i, j}^{t}(|,t')$) the probability that the right-hand side of Line~\ref{alg:SOIRETM_opr5} (\resp~\ref{alg:SOIRETM_opr6} or \ref{alg:SOIRETM_opr7}) with $\eta^t$ substituted by $t'$ evaluates to $1$. For example, $p^8_{1, 3}(\&, 10)$ denotes the probability that $g_{1, 3}^{9} \wedge g_{1, 3}^{10}$ evaluates to $1$.
Since a right-hand side may contain logical connectives $\wedge$ or $\vee$, we apply the transformations given in Table~\ref{tab:prop_diff} to estimate the ultimate probability that the right-hand side evaluates to $1$, where the special transformation is only used in the third term of Line~\ref{alg:SOIRETM_opr3} and Line~\ref{alg:SOIRETM_opr5} as well as the second term of Line~\ref{alg:SOIRETM_opr4} since these terms may have a large number of operands, while anywhere else the general transformations are used.
With the probabilities that the right-hand sides evaluate to $1$, the probability that $g_{i, j}^t$ evaluates to $1$ can be defined recursively by Equation~\ref{equ:gijt}, where we reuse $g_{i, j}^t$ to represent such a probability.
\begin{align}\label{equ:gijt}
& g_{i, j}^t= \sum_{a \in \Sigma} w^t_a \cdot 1[filter(s_{i, j}, {a})=a] \nonumber\\
& + \sum_{o \in \{?, *, +\}} w^t_o p^t_{i, j}(o) + \sum_{o \in \{\cdot, \&, |\}} w^t_o \sum_{t'=t+2}^{T} u^t_{t'} p^t_{i, j}(o, t')
\end{align}

\begin{table}[htbp]
\centering
\scalebox{0.8}{
\begin{tabular}{|c|c|c|}
    \hline
    Logical operation & General transformation & Special transformation \\
    \hline
    $A \wedge B$ & $\min(p_A, p_B)$ & - \\
    \hline
    $A \vee B$ & $\sigma_{01}(p_A+p_B)$ & $\max(p_A, p_B)$ \\
    \hline
\end{tabular}}
\caption{The transformation from logical operations to numerical computations, where $A$ and $B$ are formulae, and $p_A$ (\resp~$p_B$) is the probability that $A$ (\resp~$B$) evaluates to $1$.}
\label{tab:prop_diff}
\end{table}

For converting the return value of Algorithm~\ref{alg:SOIRETM}, we consider Line~\ref{alg:SOIRETM_alpha} and Line~\ref{alg:SOIRETM_return} in Algorithm~\ref{alg:SOIRETM} and compute the return value by Equation~\ref{equ:return}, where the second term in Equation~\ref{equ:return} ensures that all symbols occurring in $s$ appear in the target SOIRE too.
\begin{equation}\label{equ:return}
\hat{y}=g^1_{1, |s|}-\max_{a \in \Sigma} \sigma_{01}(1[a \in s]-\rho^1_a)
\end{equation}

The converted neural network is trained to minimize the objective function $\frac{1}{2}(\hat{y}-y)^2$, where $y \in \{0, 1\}$ is the ground-truth label for $r$ matching $s$.

\subsection{Faithful Encoding}

We simply refer to an assignment of the set of trainable parameters of the converted neural network as an \emph{encoding} of SOIREs. We find that encodings can one-to-one correspond to prefix notations of SOIREs for a bounded size, when it satisfies certain conditions given by Definition~\ref{def:faithful}.

\begin{definition}[Faithful encoding]\label{def:faithful}
An encoding $\theta=(w, u)$ of SOIREs with length $T$ is said to be \emph{faithful} if it satisfies all the following conditions:
\begin{enumerate}
    \item $\forall 1 \leq t \leq T, w^t$ is a one-hot vector. \label{def:faithful_1}
    \item $\forall 1 \leq t \leq T, u^t$ is either a one-hot vector or an all-zero vector. \label{def:faithful_2}
    \item $\forall 1\leq t \leq T, \sum_{t'=t+2}^{T} u^t_{t'}+\sum_{a \in \Sigma \cup \{?, *, +, \text{none}\}} w^t_a = 1$. \label{def:faithful_3}
    \item $\forall 1 \leq t \leq T-1, w^{t+1}_{\text{none}}-w^t_{\text{none}} \geq 0$. \label{def:faithful_4}
    \item $\forall 2 \leq t \leq T, \sum_{a \in \{?, +, *, \cdot, \&, |\}} w^{t-1}_a+\sum_{t'=1}^{t-2} u^{t'}_t+w^t_{\text{none}}=1$. \label{def:faithful_5}
    \item $\forall 3 \leq t \leq T, \forall 1 \leq p \leq t-2, (t-1-p)u^p_t+\sum_{p'=p+1}^{t-1}\sum_{t'=t+1}^{T}u^{p'}_{t'} \leq t-1-p$. \label{def:faithful_6}
    \item $\forall a\in \Sigma, \sum_{t=1}^{T} w^t_a \leq 1$. \label{def:faithful_7}
\end{enumerate}
\end{definition}

All conditions in a faithful encoding are translated from the construction constraints of a syntax tree.
Condition~\ref{def:faithful_1} guarantees that each vertex in the syntax tree represents a symbol, an ordinary operator or the none operator.
Condition~\ref{def:faithful_2} guarantees that each vertex has at most one right child.
Condition~\ref{def:faithful_3} guarantees that each vertex either has a right child, or represents a symbol, a unary operator or the none operator.
Condition~\ref{def:faithful_4} guarantees that if vertex $t$ represents the none operator, then vertex $t+1$ is also the none operator.
Condition~\ref{def:faithful_5} guarantees that if vertex $t$ represents the none operator, then vertex $t$ is not the child of any vertex; otherwise, vertex $t$ is the child of exactly one vertex.
Condition~\ref{def:faithful_6} guarantees that the vertices are numbered in the order of preorder traversal.
Condition~\ref{def:faithful_7} guarantees that each symbol in $\Sigma$ occurs at most once in the syntax tree. 
Example~\ref{exam:encoding} shows two faithful encodings with lengths $6$ and $8$, respectively.

By $\encre(\theta)$ we denote the prefix notation of the SOIRE interpreted from a faithful encoding $\theta$.
$\encre(\theta)$ decodes $w^t$ and $u^{t'}_t$ ($1\leq t'\leq t-2$) into the syntax tree of $r$ from $t=1$ to $T$ progressively until $w^t_{\text{none}}=1$, and then translates the constructed syntax tree of $r$ to the prefix notation of $r$.
Take Example~\ref{exam:encoding} for instance,
$\encre(\theta)=\cdot \& a b * c$ is decoded from the faithful encoding $\theta$ with length $T=8$.

Proposition~\ref{thm:img_parser} shows that $\encre(\theta)$ always yields the prefix notation of a SOIRE.

\begin{prop}\label{thm:img_parser}
For any faithful encoding $\theta$, there exists
a SOIRE $r$ such that $\encre(\theta)=\preorder(r)$.
\end{prop}

Proposition~\ref{thm:surjective}  and Proposition~\ref{thm:injection} show that $\encre(\theta)$ is surjective and injective, respectively, for a bounded size.

\begin{prop}\label{thm:surjective}
Given a bounded size $T \in \mathbb{Z}^+$,
for any SOIRE $r$ such that $|r| \leq T$, there exists a faithful encoding $\theta$ with length $T$ such that 
$\encre(\theta)=\preorder(r)$. 
\end{prop}

\begin{prop}\label{thm:injection}
Given a bounded size $T \in \mathbb{Z}^+$,
for any two different faithful encodings $\theta_1$ and $\theta_2$ with length $T$, we have $\encre(\theta_1) \neq \encre(\theta_2)$.
\end{prop}

Since $\encre(\theta)$ is both injective and surjective, faithful encodings are one-to-one corresponding to the prefix notations of SOIREs for a bounded size.

\begin{theorem}\label{thm:onetoone}
Given a bounded size $T \in \mathbb{Z}^+$, prefix notations of SOIREs $r$ with $|r|\leq T$ and faithful encodings $\theta$ with length $T$ have a one-to-one correspondence, \ie, $\encre(\theta)=\preorder(r)$. 
\end{theorem}

To make an encoding more faithful, we add one regularization for each condition to the objective function\footnote{Details of regularizations are provided in Appendix C.}.

\begin{algorithm}[htbp]
\caption{SOIRE Interpretation.}\label{alg:SOIREinterpret}
\textbf{Input}: A training set $(\Pi_{\text{train}}^+, \Pi_{\text{train}}^-)$
on the alphabet $\Sigma$, an encoding $\theta=(w, u)$ and the the beam width $\beta$.\\
\textbf{Output}: The infix notation of the target SOIRE.

\begin{algorithmic}[1]

\State Let $T$ be the first dimension of $w$;
\State Let $C^t$ be the set of candidate solutions $(r, e)$ of the subtree with the root $t$, where $r$ is the infix notation of a SOIRE and $e$ is its score;

\For{$t$ from $T$ down to $1$}
    \ForAll{$a \in \Sigma$}
    \State Add $(a, w^t_a)$ into $C^t$;
    \EndFor

    \ForAll{$(r_i, e_i) \in C^{t+1}$}
        \State Add $(r_i^?, e_i w^t_?)$ into $C^t$;
        \State Add $(r_i^*, e_i w^t_*)$ into $C^t$;
        \State Add $(r_i^+, e_i w^t_+)$ into $C^t$;
    \EndFor

    \For{$t'$ from $t+2$ to $T$}
    \ForAll{$((r_i, e_i), (r_j, e_j)) \in C^{t+1} \times C^{t'}$}
        \If{$\alpha(r_i) \cap \alpha(r_j)=\emptyset$}
        \State Add $((r_i)\cdot(r_j), e_i e_j w^t_{\cdot})$ into $C^t$;
        \State Add $((r_i)\&(r_j), e_i e_j w^t_{\&})$ into $C^t$;
        \State Add $((r_i)|(r_j), e_i e_j w^t_{|})$ into $C^t$;
        \EndIf
    \EndFor
    \EndFor
    \State Sort all $(r, e)$ in $C^t$ according to the descending order of ${e}^\frac{1}{|r|}$ and keep only the top-$\beta$ elements;
\EndFor

\State Get the accuracy of $r$ on $(\Pi_{\text{train}}^+, \Pi_{\text{train}}^-)$ for all $r$ in $C^1$;
\State \textbf{Return} $r$ in $C^1$ that has the highest accuracy;

\end{algorithmic}
\end{algorithm}

\subsection{SOIRE Interpretation}
We apply beam search to find a faithful encoding nearby the learnt encoding and then interpret it to the target SOIRE.
The algorithm is shown in Algorithm~\ref{alg:SOIREinterpret}. 
The interpretation steps are conducted from bottom to top of the syntax tree. 
We keep $\beta$ candidate SOIREs for each subtree according to the score of each candidate SOIRE, which is defined as the geometric mean of the probabilities of all operators and symbols. 
At each step, we select different operators and candidate SOIREs from the left child and right child (if any) and merge them to generate new candidates. 
At last, we calculate the accuracy of each SOIRE in the last step on the training set and pick out the SOIRE with the highest accuracy.

\section{Evaluation}
\label{experiment}

We conduct experiments to evaluate the performance of \SOIREDL on both noise-free data and noisy data.

\noindent {\bf Datasets.} 
We extract $30$ SOIREs from the RE database built by \citeauthor{Yeting2018Practical}~\shortcite{Yeting2018Practical} and generate datasets with noise from them.
The SOIREs are randomly chosen from different classes: SIRE, ICHARE, RSOIRE, SOIRE. 
We set $\Sigma$ as $10$ letters.
For each SOIRE $r$, we generate a dataset $(\Pi^+, \Pi^-)$ randomly, making sure that for all $s \in \Pi^{-}$, there exists $s'$ such that $r \models s'$ and $s$ can be modified to $s'$ by deleting a character, inserting a character at any position, replacing a character with another one, or moving a character to any other position.
For example, string $abc$ can be modified to $ac$, $abac$, $acc$ or $bca$.
We set the maximum length of strings to $20$ 
in the dataset.
For training sets and test sets, we set $|\Pi^+|=|\Pi^-|=250$, whereas for validation sets, we set $|\Pi^+|=|\Pi^-|=50$.
The noise levels are set to $\delta=\{0, 0.05, 0.1, 0.15, 0.2\}$, where for each $\delta$, we reverse the labels for $|\Pi^+|\delta$ positive strings and $|\Pi^-|\delta$ negative strings in the training and validation sets.

\noindent {\bf Competitors.} 
We choose approaches \iSOIRE~\cite{Yeting2019effective}, \GenICHARE~\cite{Xiaolan2018Interleaving}, \iSIRE~\cite{Yeting2020Negative} and \RERNN~\cite{Chengyue2020Cold} as competitors. 
\iSOIRE learns RSOIREs and \GenICHARE learns ICHAREs from positive strings only. 
\iSIRE learns SIREs from both positive and negative strings. 
\RERNN embeds a weighted FSA to improve the performance on text classification. It can also learn an automaton if we randomly initialize the parameters.
Thus we also compare the performance between \SOIREDL and \RERNN. 

\noindent {\bf Settings.} 
We implement \iSOIRE, \GenICHARE, \iSIRE according to their papers, and reuse the source code of \RERNN. The hyper-parameters of \RERNN are set as default, except that the number of states is set to $100$ and the threshold in interpretation to $0.12$ for achieving the best accuracy.
We train \SOIREDL with the AdamW optimizer. 
The hyper-parameters of \SOIREDL are set as follows: the bounded size $T$ is $4|\Sigma|-2$ according to Proposition~\ref{thm:T_bound}, the batch size is $64$, the regularization coefficient $\lambda$ is $0$, and the beam width $\beta$ is $500$.
The optimal $\lambda$ is selected from $\{0,10^{-3},10^{-2},10^{-1},1,10\}$ and $\beta$ from $\{10,50,100,300,500,1000\}$ for maximizing the accuracy on the validation set.
\iSOIRE, \GenICHARE and \iSIRE use the union of the training and validation sets for learning.

All experiments were conducted on a Linux machine equipped with an Intel Xeon Gold 6248R processor with $126$ GB RAM and a single NVIDIA A$100$. 
%\iSIRE runs on $10$ CPU threads and \SOIREDL runs on the GPU. 
We train \SOIREDL with five learning rates $0.01, 0.05, 0.1, 0.15, 0.2$ and select the SOIRE achieving the best accuracy on the validation set.
Therefore, the running time of \SOIREDL is the sum of training time and interpretation time with five learning rates.
The time limit for each approach is set to $5000$ seconds.

\begin{prop}\label{thm:T_bound}
For any SOIRE $r$ over $\Sigma$, there exists another SOIRE $r'$ over $\Sigma$ and a faithful encoding $\theta$ with length $4|\Sigma|-2$ such that $\preorder(r')=\encre(\theta)$ and $\{s|r' \models s\}=\{s|r \models s\}$.
\end{prop}

\noindent {\bf Evaluate metrics.} 
We use accuracy on the test set to evaluate the performance of all approaches. For \SOIREDL and \RERNN, We introduce faithfulness defined as $\frac{N_=}{\left| \Pi^{+} \right| + \left| \Pi^{-} \right|}$ to evaluate the consistency between the neural network and the interpreted SOIRE (\SOIREDL) or automata (\RERNN), where $N_=$ is the number of test strings that the neural network and the SOIRE or automata predicts the same label.

\begin{table}[htbp]
\scalebox{0.58}{
\begin{tabular}{|c|ccc|ccccc|}
    \hline
    & \multicolumn{3}{c|}{Positive and negative strings} & \multicolumn{5}{c|}{Positive strings only} \\
    \cline{2-9}
    Data & iSI & \multirow{2}{*}{RE2RNN} & \multirow{2}{*}{SOIREDL} & iSO  & GenIC   & iSI &  \multirow{2}{*}{RE2RNN} & \multirow{2}{*}{SOIREDL} \\
    set &  RE   &                        &                          & IRE  & HARE & RE  &                          &  \\ \hline
    1 & 86.0 & 52.4 (94.4) & \textbf{100.0} (100.0) & \textbf{89.4} & \textbf{89.4} & 83.8 & 48.0 (50.0) & 57.0 (57.0) \\
    2 & 77.4 & 48.8 (91.4) & \textbf{100.0} (100.0) & \textbf{100.0} & \textbf{100.0} & \textbf{100.0} & 50.4 (50.0) & \textbf{100.0} (100.0) \\
    3 & 90.8 & 50.6 (77.4) & \textbf{100.0} (100.0) & \textbf{100.0} & 97.2 & 95.6 & 50.6 (49.6) & 65.4 (65.4) \\
    4 & 72.4 & 49.0 (80.2) & \textbf{99.6} (73.4) & \textbf{73.8} & 72.6 & 73.4 & 49.6 (49.8) & 61.4 (61.4) \\
    5 & \textbf{90.8} & 52.6 (91.4) & 58.2 (87.2) & \textbf{100.0} & \textbf{100.0} & 86.2 & 49.6 (50.2) & 52.8 (52.8) \\
    6 & 77.8 & 51.6 (62.2) & \textbf{93.4} (93.0) & \textbf{100.0} & \textbf{100.0} & 70.4 & 50.2 (50.0) & 52.6 (52.6) \\
    7 & 81.2 & 52.2 (95.8) & \textbf{99.2} (99.2) & \textbf{100.0} & 96.4 & 89.0 & 49.2 (49.8) & 69.4 (69.4) \\
    8 & 76.2 & 49.8 (88.0) & \textbf{100.0} (100.0) & \textbf{100.0} & \textbf{100.0} & \textbf{100.0} & 49.8 (50.0) & \textbf{100.0} (100.0) \\
    9 & 93.8 & 44.2 (48.4) & \textbf{98.4} (98.4) & \textbf{99.8} & 98.8 & 98.0 & 47.2 (49.6) & 81.6 (81.6) \\
    10 & 94.8 & 50.2 (89.8) & \textbf{99.8} (100.0) & \textbf{99.8} & \textbf{99.8} & 82.8 & 44.4 (50.2) & 61.2 (61.2) \\
    11 & \textbf{91.0} & 52.4 (88.6) & 89.2 (91.0) & \textbf{100.0} & \textbf{100.0} & 91.4 & 47.6 (50.2) & 57.4 (57.4) \\
    12 & 78.6 & 51.2 (98.4) & \textbf{100.0} (100.0) & \textbf{100.0} & \textbf{100.0} & \textbf{100.0} & 50.8 (49.4) & \textbf{100.0} (100.0) \\
    13 & 96.6 & 52.8 (50.2) & \textbf{100.0} (100.0) & \textbf{100.0} & \textbf{100.0} & 83.6 & 51.2 (49.6) & 67.0 (67.0) \\
    14 & 74.4 & 50.0 (79.0) & \textbf{84.8} (71.6) & 70.0 & \textbf{74.4} & 68.8 & 49.0 (50.0) & 54.4 (54.4) \\
    15 & 95.2 & 54.0 (49.8) & \textbf{96.2} (100.0) & \textbf{100.0} & \textbf{100.0} & \textbf{100.0} & 47.8 (50.2) & 66.2 (66.2) \\
    16 & \textbf{96.8} & 49.0 (71.4) & 94.8 (100.0) & \textbf{100.0} & \textbf{100.0} & 75.4 & 49.4 (50.0) & 75.4 (75.4) \\
    17 & 91.0 & 46.2 (87.2) & \textbf{100.0} (100.0) & \textbf{100.0} & \textbf{100.0} & \textbf{100.0} & 50.6 (50.0) & \textbf{100.0} (100.0) \\
    18 & 81.0 & 42.8 (78.8) & \textbf{87.4} (99.2) & 87.4 & \textbf{100.0} & 82.0 & 40.2 (50.2) & 53.0 (53.0) \\
    19 & 88.2 & 50.0 (63.8) & \textbf{100.0} (93.4) & \textbf{100.0} & \textbf{100.0} & 88.0 & 50.4 (50.0) & 54.6 (54.6) \\
    20 & \textbf{93.4} & 45.2 (54.4) & 83.4 (90.0) & \textbf{100.0} & 99.2 & 95.8 & 47.4 (50.0) & 56.0 (51.2) \\
    21 & 69.8 & 48.8 (96.2) & \textbf{100.0} (100.0) & \textbf{71.2} & \textbf{71.2} & \textbf{71.2} & 49.8 (50.4) & \textbf{71.2} (71.2) \\
    22 & 90.6 & 47.6 (50.6) & \textbf{100.0} (100.0) & \textbf{100.0} & \textbf{100.0} & 91.4 & 50.0 (50.2) & 58.2 (58.2) \\
    23 & 85.6 & 32.8 (84.4) & \textbf{86.8} (65.2) & \textbf{90.0} & \textbf{90.0} & 88.0 & 50.0 (48.8) & 56.8 (56.8) \\
    24 & 69.4 & 49.0 (57.2) & \textbf{74.6} (76.8) & \textbf{77.6} & 76.0 & 71.4 & 47.6 (50.0) & 54.2 (54.2) \\
    25 & 67.8 & 54.2 (93.4) & \textbf{99.8} (74.6) & \textbf{70.0} & \textbf{70.0} & \textbf{70.0} & 50.2 (50.0) & 69.6 (69.6) \\
    26 & \textbf{80.2} & 50.4 (50.4) & 63.8 (98.2) & \textbf{100.0} & \textbf{100.0} & 85.2 & 51.0 (49.6) & 63.8 (63.8) \\
    27 & 92.0 & 52.0 (91.6) & \textbf{96.4} (96.4) & \textbf{100.0} & \textbf{100.0} & 97.4 & 52.2 (50.2) & 67.6 (67.6) \\
    28 & 65.0 & 54.8 (95.2) & \textbf{100.0} (98.6) & \textbf{65.6} & \textbf{65.6} & \textbf{65.6} & 50.0 (50.0) & \textbf{65.6} (65.6) \\
    29 & 93.4 & 56.2 (75.8) & \textbf{97.6} (97.2) & \textbf{100.0} & 99.8 & 81.2 & 49.2 (49.8) & 54.2 (54.2) \\
    30 & 60.4 & 41.4 (61.6) & \textbf{78.8} (79.6) & \textbf{61.0} & \textbf{61.0} & 60.4 & 49.8 (49.8) & 54.8 (54.8) \\
    \hline Avg. & 83.4 & 49.4 (76.6) & \textbf{92.7} (92.8) & 91.9 & \textbf{92.0} & 84.9 & 49.1 (49.9) & 66.7 (66.6) \\\hline 
    \end{tabular}}
    \caption{Accuracy (\%) on noise-free data with best results in bold. %`Avg.' means average accuracy.
    For X(Y), X denotes the accuracy of the learnt SOIRE or automaton, and Y the accuracy of the neural network.}
    \label{tab:noise_free}
\end{table}

\noindent {\bf Comparison on noise-free data.} 
The results of different approaches on noise-free data are shown in Table~\ref{tab:noise_free}. 
For learning from both positive and negative strings, \SOIREDL outperforms \iSIRE and \RERNN on almost all datasets, achieves comparable performance with \iSOIRE and \GenICHARE (both of which learn from positive strings only), and achieves the highest average accuracy among all approaches. Regarding the accuracy of the intermediate neural network, \SOIREDL is also superior to \RERNN. 
These results show that \SOIREDL achieves the SOTA performance on noise-free data.

\begin{figure}[ht]
\centering
\includegraphics[width=0.45\textwidth]{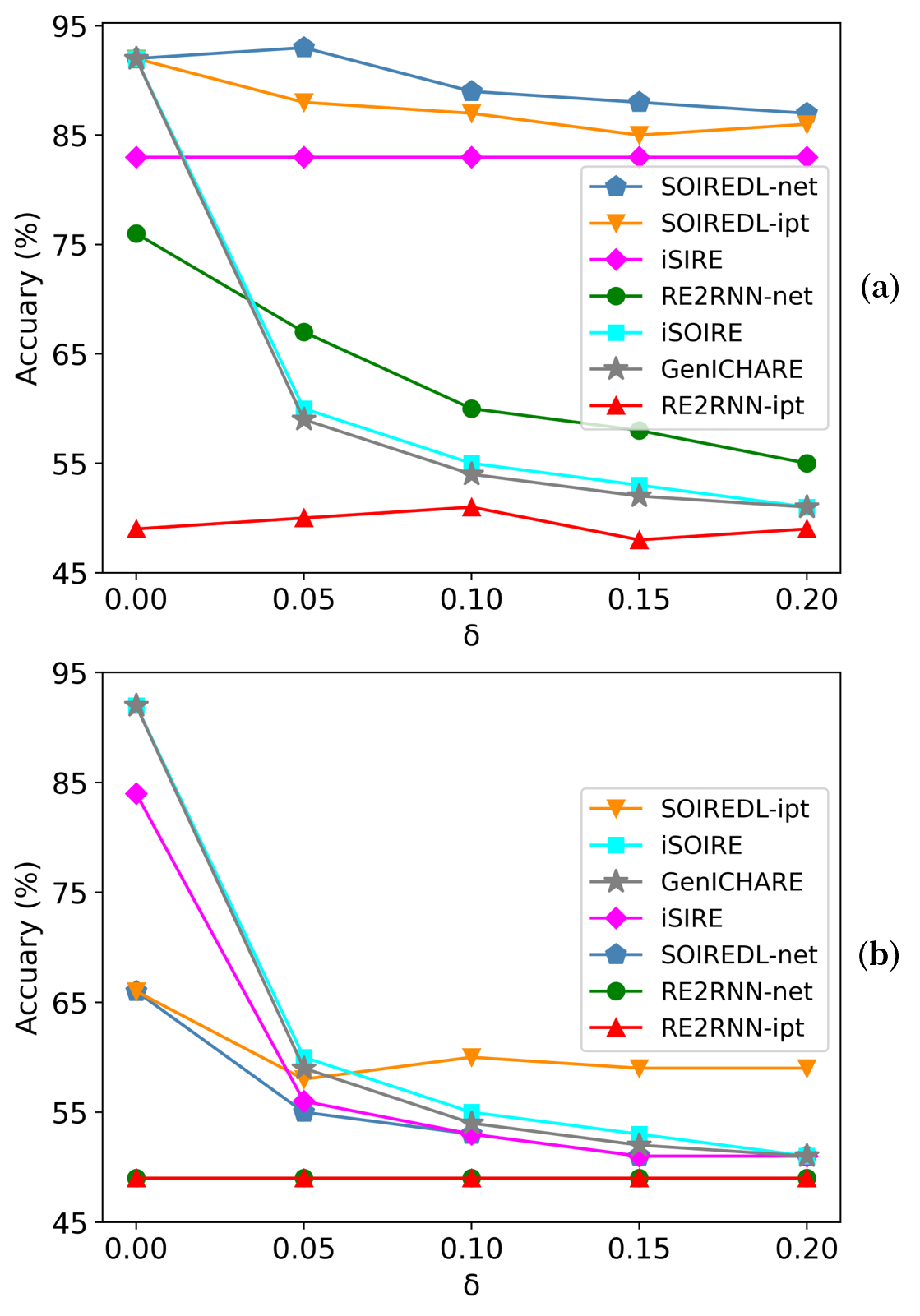}
\caption{Average accuracy (\%) on test sets at different noise levels $\delta$. 
            SOIREDL-ipt and RE2RNN-ipt represent the learnt SOIREs or automata, whereas SOIREDL-net and RE2RNN-net represent the neural networks.
            (a) Positive and negative strings. 
            (b) Positive strings only.}
\label{fig:noise}
\end{figure}

\noindent {\bf Comparison on noisy data.} 
The average accuracy of different approaches on noisy data is shown in Figure~\ref{fig:noise} (a).
The performance of \iSOIRE and \GenICHARE drops sharply when noise is present, suggesting that they are not robust on noisy data. 
The average accuracy of \RERNN also decreases quickly when the noise level increases, and so does the neural network of it. 
Both \SOIREDL and \iSIRE perform well on noisy data. 
The average accuracy of \SOIREDL slightly decreases when the noise level increases, 
but it still keeps higher than others at all noise levels. This suggests that \SOIREDL is the most robust on noisy data.

\begin{figure}[htbp]
\centering
\includegraphics[width=0.42\textwidth]{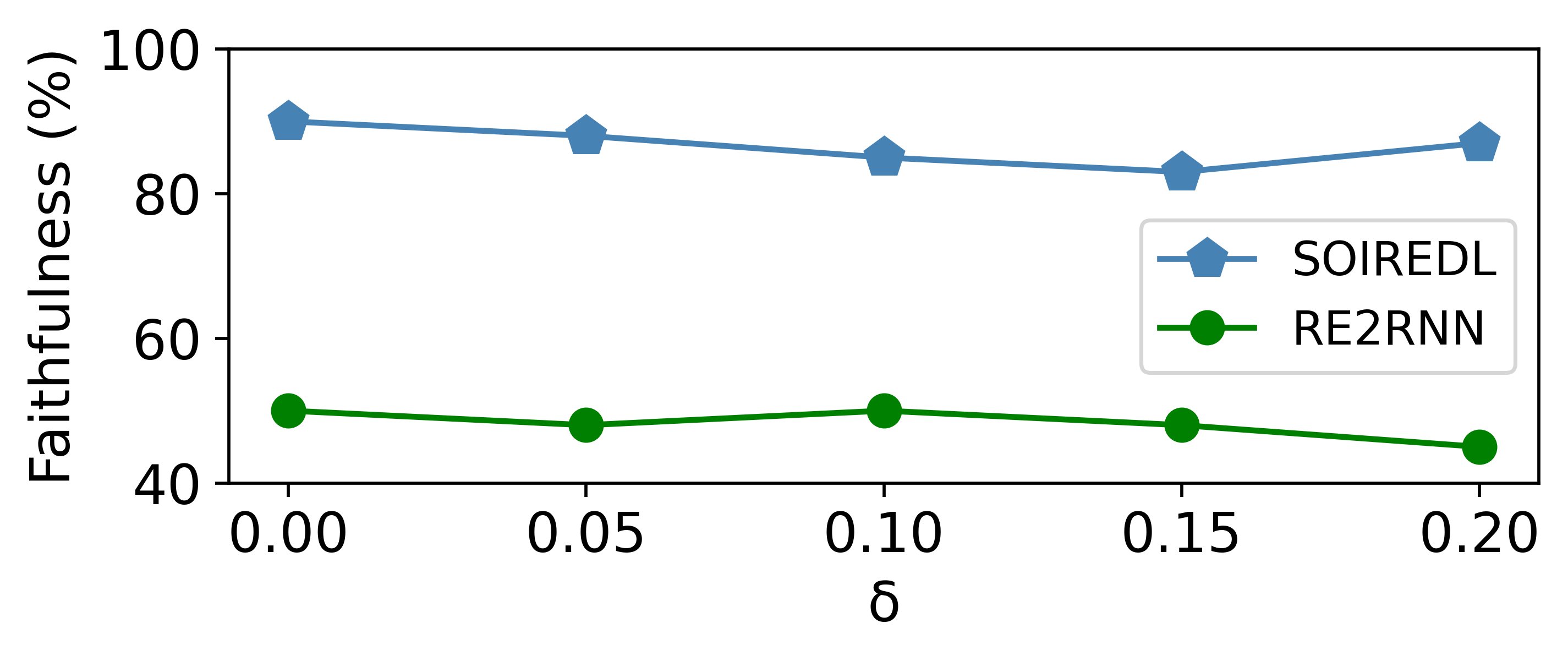}
\caption{Average faithfulness (\%) of \SOIREDL and \RERNN on test sets at different noise levels $\delta$.}
\label{fig:faithful}
\end{figure}

\noindent {\bf Comparison in terms of faithfulness.}
The average faithfulness of \SOIREDL and \RERNN is shown in Figure~\ref{fig:faithful}. 
Obviously, the faithfulness of \SOIREDL is much higher than that of \RERNN and keeps over $80\%$ at all noise levels, suggesting that the neural network of \SOIREDL and its interpreted SOIRE are more consistent in performing SOIRE matching.

\noindent {\bf Case study.} 
We pick one RE from each subclass of SOIREs considered in our experiments to show their learnt results, reported in Table~\ref{tab:case_study}. 
Although the ground-truth REs and the SOIREs learnt by \SOIREDL are not exactly the same in the subclasses ICHARE and RSOIRE, they still belong to the same subclass. These results show that \SOIREDL is able to learn different subclasses of SOIREs.

\begin{table}[htbp]
\centering
\scalebox{0.80}{
    \begin{tabular}{|c|c|c|c|c|}
    \hline
    Subclass     & Dataset & Ground-truth SOIREs                                 & SOIREDL                                            \\
    \hline
    SIRE      & 13      & $a^{?} \& b^{*} \& c^{?}$                     & $a^{?} \& c^{?} \& b^{*}$                           \\
    ICHARE & 22      & $\left( a|b \right)^{*}c^{*}d^{*}$            & $\left( a^{*} \& b^{*} \right)c^{*}d^{*}$           \\
    RSOIRE    & 3       & $a^{+} | \left( b|c \right)^{*} | d^{+}$      & $\left( b^{+}|c \right)^{*} | a^{*} | d^{*}$        \\
    SOIRE     & 1       & $\left( \left( a|b \right)c^{*} \right)^{+}d$ & $\left( \left( a|b \right)c^{*} \right)^{+}d$       \\
    \hline
\end{tabular}}
    \caption{The ground-truth SOIREs and the SOIREs learnt by SOIREDL on different subclasses of SOIREs.}
    \label{tab:case_study}
\end{table}

\begin{figure}[htbp]
\centering
\includegraphics[width=0.42\textwidth]{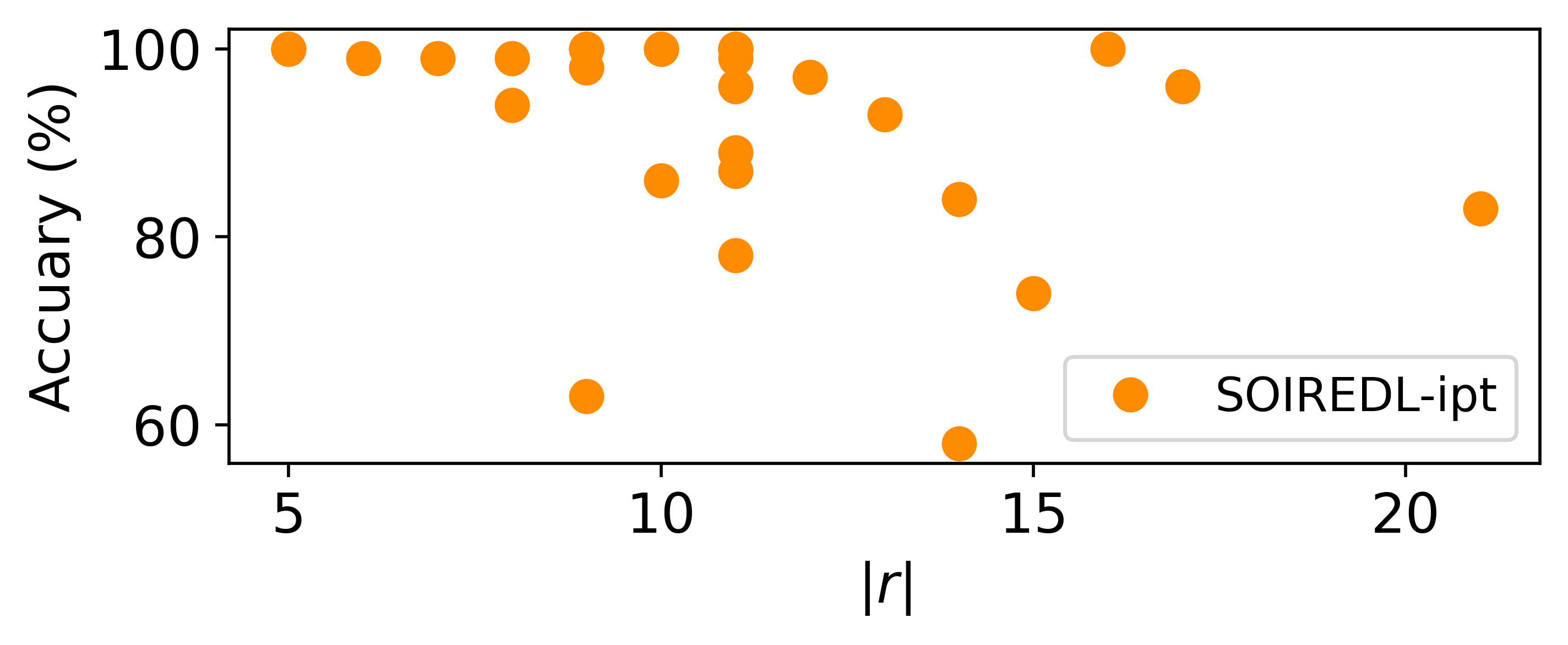}
\caption{The relation of the accuracy (\%) of a  SOIRE learnt by \SOIREDL and the size $|r|$ of the ground-truth SOIRE $r$.}
\label{fig:size-acc}
\end{figure}

We also analyse the relation between the accuracy of the SOIRE learnt by \SOIREDL and the size of the ground-truth SOIRE. 
Figure~\ref{fig:size-acc} shows that the accuracy decreases when the size of the ground-truth SOIRE increases.
This may be due to the difficulty for a neural network to capture the long-distance dependency in SOIRE matching.

\noindent {\bf The necessity for using negative strings.}
The results of learning from positive strings only are shown in Table~\ref{tab:noise_free} and Figure~\ref{fig:noise} (b). 
Neural networks do not perform well because they prefer to classify unseen strings as positive ones when training on positive strings only.
Even worse, when noise is present, the accuracy of any competitor drops sharply to no more than $60\%$ as shown in Figure~\ref{fig:noise} (b).
%All approaches cannot work well with noisy data when learning from positive strings only.
This suggests that negative strings are crucial in effectively learning with noisy data, possibly because they infer what kinds of strings that the target SOIRE cannot match.
By learning from positive and negative strings, both \SOIREDL and \RERNN achieve significantly better performance; in particular, \SOIREDL outperforms \iSOIRE and \GenICHARE in terms of average accuracy, as shown in Table~\ref{tab:noise_free} and Figure~\ref{fig:noise} (a).

\section{Conclusion and Future Work}
\label{conclusion}

In this paper, we have proposed a noise-tolerant differentiable learning approach \SOIREDL and a matching algorithm \SOIRETM based on the syntax tree for SOIREs. 
The neural network introduced in \SOIREDL simulates the matching process of \SOIRETM. 
Theoretically, the faithful encodings learnt by \SOIREDL one-to-one correspond to SOIREs for a bounded size.
Experimental results have demonstrated higher performance compared with the SOTA approaches.
In the future work, we will tackle the problem of long dependency in SOIRE matching and extend our approach to other subclasses of REs.

\section{Acknowledgments}

We thank Kunxun Qi for discussion on the paper and anonymous referees for helpful comments.

This paper was supported by the National Natural Science Foundation of China (No. 62276284, 61976232, 61876204), the National Key Research and Development Program of China (No. 2021YFA1000504), Guangdong Basic and Applied Basic Research Foundation (No. 2022A1515011355), Guangzhou Science and Technology Project (No. 202201011699), Guizhou Science Support Project (No. 2022-259), as well as Humanities and Social Science Research Project of Ministry of Education (No. 18YJCZH006).

\bibliography{SOIRE}

\begin{thebibliography}{39}
\providecommand{\natexlab}[1]{#1}

\bibitem[{Bex et~al.(2006)Bex, Neven, Schwentick, and Tuyls}]{Bex2006}
Bex, G.~J.; Neven, F.; Schwentick, T.; and Tuyls, K. 2006.
\newblock Inference of Concise DTDs from {XML} Data.
\newblock In \emph{VLDB}, 115--126.

\bibitem[{Bojanczyk et~al.(2006)Bojanczyk, Muscholl, Schwentick, Segoufin, and
  David}]{Mikolaj2006Two}
Bojanczyk, M.; Muscholl, A.; Schwentick, T.; Segoufin, L.; and David, C. 2006.
\newblock Two-Variable Logic on Words with Data.
\newblock In \emph{LICS}, 7--16.

\bibitem[{Cohen, Yang, and Mazaitis(2020)}]{William2020TensorLog}
Cohen, W.~W.; Yang, F.; and Mazaitis, K. 2020.
\newblock TensorLog: {A} Probabilistic Database Implemented Using Deep-Learning
  Infrastructure.
\newblock \emph{J. Artif. Intell. Res.}, 67: 285--325.

\bibitem[{Colazzo et~al.(2013)Colazzo, Ghelli, Pardini, and
  Sartiani}]{Colazzo2013Efficient}
Colazzo, D.; Ghelli, G.; Pardini, L.; and Sartiani, C. 2013.
\newblock Efficient asymmetric inclusion of regular expressions with
  interleaving and counting for {XML} type-checking.
\newblock \emph{Theor. Comput. Sci.}, 492: 88--116.

\bibitem[{Freydenberger and K{\"{o}}tzing(2015)}]{Dominik2015Fast}
Freydenberger, D.~D.; and K{\"{o}}tzing, T. 2015.
\newblock Fast Learning of Restricted Regular Expressions and DTDs.
\newblock \emph{Theory Comput. Syst.}, 57(4): 1114--1158.

\bibitem[{Galassi and Giordana(2005)}]{Galassi2005Learning}
Galassi, U.; and Giordana, A. 2005.
\newblock Learning Regular Expressions from Noisy Sequences.
\newblock In \emph{SARA}, volume 3607, 92--106.

\bibitem[{Gao et~al.(2022)Gao, Inoue, Cao, and Wang}]{Kun2022Learning}
Gao, K.; Inoue, K.; Cao, Y.; and Wang, H. 2022.
\newblock Learning First-Order Rules with Differentiable Logic Program
  Semantics.
\newblock In \emph{IJCAI}, 3008--3014.

\bibitem[{Gischer(1981)}]{Gischer1981Shuffle}
Gischer, J.~L. 1981.
\newblock Shuffle Languages, Petri Nets, and Context-Sensitive Grammars.
\newblock \emph{Commun. {ACM}}, 24(9): 597--605.

\bibitem[{Huang et~al.(2021)Huang, Li, Chen, Samel, Naik, Song, and
  Si}]{Jiani2021Scallop}
Huang, J.; Li, Z.; Chen, B.; Samel, K.; Naik, M.; Song, L.; and Si, X. 2021.
\newblock Scallop: From Probabilistic Deductive Databases to Scalable
  Differentiable Reasoning.
\newblock In \emph{NeurIPS}, 25134--25145.

\bibitem[{Jiang, Jin, and Tu(2021)}]{Chengyue2021Neuralizing}
Jiang, C.; Jin, Z.; and Tu, K. 2021.
\newblock Neuralizing Regular Expressions for Slot Filling.
\newblock In \emph{EMNLP}, 9481--9498.

\bibitem[{Jiang et~al.(2020)Jiang, Zhao, Chu, Shen, and Tu}]{Chengyue2020Cold}
Jiang, C.; Zhao, Y.; Chu, S.; Shen, L.; and Tu, K. 2020.
\newblock Cold-Start and Interpretability: Turning Regular Expressions into
  Trainable Recurrent Neural Networks.
\newblock In \emph{EMNLP}, 3193--3207.

\bibitem[{Kearns and Li(1988)}]{Kearns1988Learning}
Kearns, M.~J.; and Li, M. 1988.
\newblock Learning in the Presence of Malicious Errors (Extended Abstract).
\newblock In \emph{STOC}, 267--280.

\bibitem[{Kuhlmann and Satta(2009)}]{Kuhlmann2009Treebank}
Kuhlmann, M.; and Satta, G. 2009.
\newblock Treebank Grammar Techniques for Non-Projective Dependency Parsing.
\newblock In \emph{EACL}, 478--486.

\bibitem[{Li et~al.(2020{\natexlab{a}})Li, Cao, Chen, Ge, Xu, and
  Peng}]{Yeting2020FlashSchema}
Li, Y.; Cao, J.; Chen, H.; Ge, T.; Xu, Z.; and Peng, Q. 2020{\natexlab{a}}.
\newblock FlashSchema: Achieving High Quality {XML} Schemas with Powerful
  Inference Algorithms and Large-scale Schema Data.
\newblock In \emph{ICDE}, 1962--1965.

\bibitem[{Li et~al.(2020{\natexlab{b}})Li, Chen, Zhang, Huang, and
  Zhang}]{Yeting2020Negative}
Li, Y.; Chen, H.; Zhang, L.; Huang, B.; and Zhang, J. 2020{\natexlab{b}}.
\newblock Inferring Restricted Regular Expressions with Interleaving from
  Positive and Negative Samples.
\newblock In \emph{PAKDD}, volume 12085, 769--781.

\bibitem[{Li et~al.(2019{\natexlab{a}})Li, Chen, Zhang, and
  Zhang}]{Yeting2019effective}
Li, Y.; Chen, H.; Zhang, X.; and Zhang, L. 2019{\natexlab{a}}.
\newblock An effective algorithm for learning single occurrence regular
  expressions with interleaving.
\newblock In \emph{IDEAS}, 24:1--24:10.

\bibitem[{Li et~al.(2018)Li, Chu, Mou, Dong, and Chen}]{Yeting2018Practical}
Li, Y.; Chu, X.; Mou, X.; Dong, C.; and Chen, H. 2018.
\newblock Practical Study of Deterministic Regular Expressions from Large-scale
  {XML} and Schema Data.
\newblock In \emph{IDEAS}, 45--53.

\bibitem[{Li et~al.(2021)Li, Li, Xu, Cao, Chen, Hu, Chen, and
  Cheung}]{Yeting2021TRANSREGEX}
Li, Y.; Li, S.; Xu, Z.; Cao, J.; Chen, Z.; Hu, Y.; Chen, H.; and Cheung, S.
  2021.
\newblock {TRANSREGEX:} Multi-modal Regular Expression Synthesis by
  Generate-and-Repair.
\newblock In \emph{ICSE}, 1210--1222.

\bibitem[{Li et~al.(2019{\natexlab{b}})Li, Zhang, Cao, Chen, and
  Gao}]{Yeting2019koire}
Li, Y.; Zhang, X.; Cao, J.; Chen, H.; and Gao, C. 2019{\natexlab{b}}.
\newblock Learning k-Occurrence Regular Expressions with Interleaving.
\newblock In \emph{DASFAA}, volume 11447, 70--85.

\bibitem[{Makoto and Clark(2003)}]{relaxng}
Makoto, M.; and Clark, J. 2003.
\newblock RELAX NG.
\newblock https://relaxng.org/.

\bibitem[{Martens et~al.(2017)Martens, Neven, Niewerth, and
  Schwentick}]{Wim2017BonXai}
Martens, W.; Neven, F.; Niewerth, M.; and Schwentick, T. 2017.
\newblock BonXai: Combining the Simplicity of {DTD} with the Expressiveness of
  {XML} Schema.
\newblock \emph{{ACM} Trans. Database Syst.}, 42(3): 15:1--15:42.

\bibitem[{Mayer and Stockmeyer(1994)}]{Alain1994Complexity}
Mayer, A.~J.; and Stockmeyer, L.~J. 1994.
\newblock The Complexity of Word Problems - This Time with Interleaving.
\newblock \emph{Inf. Comput.}, 115(2): 293--311.

\bibitem[{Mensch and Blondel(2018)}]{Arthur2018Differentiable}
Mensch, A.; and Blondel, M. 2018.
\newblock Differentiable Dynamic Programming for Structured Prediction and
  Attention.
\newblock In \emph{ICML}, volume~80, 3459--3468.

\bibitem[{Minervini et~al.(2020{\natexlab{a}})Minervini, Bosnjak,
  Rockt{\"{a}}schel, Riedel, and Grefenstette}]{Pasquale2020Differentiable}
Minervini, P.; Bosnjak, M.; Rockt{\"{a}}schel, T.; Riedel, S.; and
  Grefenstette, E. 2020{\natexlab{a}}.
\newblock Differentiable Reasoning on Large Knowledge Bases and Natural
  Language.
\newblock In \emph{AAAI}, 5182--5190.

\bibitem[{Minervini et~al.(2020{\natexlab{b}})Minervini, Riedel, Stenetorp,
  Grefenstette, and Rockt{\"{a}}schel}]{Pasquale2020Learning}
Minervini, P.; Riedel, S.; Stenetorp, P.; Grefenstette, E.; and
  Rockt{\"{a}}schel, T. 2020{\natexlab{b}}.
\newblock Learning Reasoning Strategies in End-to-End Differentiable Proving.
\newblock In \emph{ICML}, volume 119, 6938--6949.

\bibitem[{Nivre(2009)}]{Joakim2009Projective}
Nivre, J. 2009.
\newblock Non-Projective Dependency Parsing in Expected Linear Time.
\newblock In \emph{ACL}, 351--359.

\bibitem[{Rockt{\"{a}}schel and Riedel(2017)}]{Tim2017Proving}
Rockt{\"{a}}schel, T.; and Riedel, S. 2017.
\newblock End-to-end Differentiable Proving.
\newblock In \emph{NeurIPS}, 3788--3800.

\bibitem[{Sadeghian et~al.(2019)Sadeghian, Armandpour, Ding, and
  Wang}]{Ali2019DRUM}
Sadeghian, A.; Armandpour, M.; Ding, P.; and Wang, D.~Z. 2019.
\newblock {DRUM:} End-To-End Differentiable Rule Mining On Knowledge Graphs.
\newblock In \emph{NeurIPS}, 15321--15331.

\bibitem[{Wang et~al.(2019)Wang, Donti, Wilder, and Kolter}]{Wang2019SATNet}
Wang, P.; Donti, P.~L.; Wilder, B.; and Kolter, J.~Z. 2019.
\newblock SATNet: Bridging deep learning and logical reasoning using a
  differentiable satisfiability solver.
\newblock In \emph{ICML}, volume~97, 6545--6554.

\bibitem[{Wang et~al.(2020)Wang, Stepanova, Domokos, and
  Kolter}]{Wang2020numerical}
Wang, P.; Stepanova, D.; Domokos, C.; and Kolter, J.~Z. 2020.
\newblock Differentiable learning of numerical rules in knowledge graphs.
\newblock In \emph{ICLR}.

\bibitem[{Wang(2021{\natexlab{a}})}]{Xiaofan2021UnorderCounting}
Wang, X. 2021{\natexlab{a}}.
\newblock Inferring Deterministic Regular Expression with Unorder and Counting.
\newblock In \emph{DASFAA}, volume 12682, 235--252.

\bibitem[{Wang(2021{\natexlab{b}})}]{Xiaofan2021Shuffle}
Wang, X. 2021{\natexlab{b}}.
\newblock Learning Finite Automata with Shuffle.
\newblock In \emph{PAKDD}, volume 12713, 308--320.

\bibitem[{Wang(2022)}]{Xiaofan2022Membership}
Wang, X. 2022.
\newblock Membership Algorithm for Single-Occurrence Regular Expressions with
  Shuffle and Counting.
\newblock In \emph{DASFAA}, volume 13245, 526--542.

\bibitem[{Wang and Chen(2018)}]{Xiaofan2018Counting}
Wang, X.; and Chen, H. 2018.
\newblock Inferring Deterministic Regular Expression with Counting.
\newblock In \emph{Conceptual Modeling - 37th International Conference, {ER}},
  volume 11157, 184--199.

\bibitem[{Wang and Chen(2020)}]{Xiaofan2020Unorder}
Wang, X.; and Chen, H. 2020.
\newblock Inferring Deterministic Regular Expression with Unorder.
\newblock In \emph{SOFSEM}, volume 12011, 325--337.

\bibitem[{Wang and Zhang(2021)}]{Xiaofan2021Discovering}
Wang, X.; and Zhang, X. 2021.
\newblock Discovering an Algorithm Actually Learning Restricted Single
  Occurrence Regular Expression with Interleaving.
\newblock \emph{CoRR}, abs/2103.10546.

\bibitem[{Yang, Yang, and Cohen(2017)}]{Fan2017Differentiable}
Yang, F.; Yang, Z.; and Cohen, W.~W. 2017.
\newblock Differentiable Learning of Logical Rules for Knowledge Base
  Reasoning.
\newblock In \emph{NeurIPS}, 2319--2328.

\bibitem[{Yang and Song(2020)}]{Yuan2020Learn}
Yang, Y.; and Song, L. 2020.
\newblock Learn to Explain Efficiently via Neural Logic Inductive Learning.
\newblock In \emph{ICLR}.

\bibitem[{Zhang et~al.(2018)Zhang, Li, Cui, Dong, and
  Chen}]{Xiaolan2018Interleaving}
Zhang, X.; Li, Y.; Cui, F.; Dong, C.; and Chen, H. 2018.
\newblock Inference of a Concise Regular Expression Considering Interleaving
  from {XML} Documents.
\newblock In \emph{PAKDD}, volume 10938, 389--401.

\end{thebibliography}

\newpage

\section{Appendix A}
\label{app:alg}

\subsection{From prefix notation to the infix notation of a SOIRE}

Algorithm~\ref{alg:prefix2middle} is a function to obtain the infix notaiton of a SOIRE from the prefix notaiton.
  
\begin{algorithm}[h]
\caption{From prefix notation to the infix notation of a SOIRE.}\label{alg:prefix2middle}
\textbf{Input}: The prefix notation of the SOIRE $r_p$.\\
\textbf{Output}: The infix notation of the SOIRE $r_i$.

\begin{algorithmic}[1]

\State Let $\Gamma$ be an empty stack;

\For{$i \gets |r_p|$ downto $1$}
    \State Let $o^i$ be the $i$ th character of $r_p$;
    \If{$o^i \in \Sigma$}
    \State Push $(o^i)$ into stack $\Gamma$;
    \EndIf
    \If{$o^i \in \{?, *, +\}$}
    \State Pop an element from stack $\Gamma$, denoted as $r$;
    \State Push $\text{concat}(`(', r, o^i, `)')$ into stack $\Gamma$;
    \EndIf
    \If{$o^i \in \{\cdot, \&, |\}$}
    \State Pop an element from stack $\Gamma$, denoted as $r_2$;
    \State Pop an element from stack $\Gamma$, denoted as $r_1$;
    \State Push $\text{concat}(`(', r_1, o^i, r_2, `)')$ into stack $\Gamma$;
    \EndIf
\EndFor

\State Pop an element from stack $\Gamma$, denoted as $r_i$;
\State \textbf{Return} $r_i$;

\end{algorithmic}

\end{algorithm}

\subsection{Faithful encoding interpreter}

\begin{algorithm}[htbp]
    \caption{Faithful encoding interpreter \encre.}\label{alg:faithful_interpret}
    \textbf{Input}: The faithful encoding of an SOIRE, $\theta=(w, u)$.\\
    \textbf{Output}: The prefix notation of an SOIRE $r$.
  
    \begin{algorithmic}[1]
  
    \State Let $r$ be a empty string;
    \State Let $T$ be bounded size of $r$;
  
    \For{$t \gets 1$ to $T$}
      \If{$w^t_{\text{none}}=1$}
        \State \textbf{\bf Break};
      \EndIf
      \ForAll{$a \in \mathbb{B} \setminus \{\text{none}\}$}
        \If{$w^t_a=1$}
          \State $r \gets \text{concat}(a, r)$;
        \EndIf
      \EndFor
    \EndFor
  
    \State \textbf{Return} $r$;
  
    \end{algorithmic}
  \end{algorithm}

\section{Appendix B}
\label{app:proof}

\subsection{Proof of Theorem~\ref{thm:match_cond}}

\noindent {\bf Proof}:

\begin{enumerate}
  \item If $r \models s$, then $filter(s, \alpha(r))=s$ and $r \models filter(s, \alpha(r))$.

  Because $r \models s$, $\alpha(r)$ contains all the symbols in $s$ and $filter(s, \alpha(r))=s$ (Definition of SOIRE matching). 
  Because $r \models s$ and $filter(s, \alpha(r))=s$, $r \models filter(s, \alpha(r))$. 

  \item If $filter(s, \alpha(r))=s$ and $r \models filter(s, \alpha(r))$, than $r \models s$.

  Because $r \models filter(s, \alpha(r))$ and $filter(s, \alpha(r))=s$, $r \models s$. 

\end{enumerate}

Theorem~\ref{thm:match_cond} has been proved.

\subsection{Proof of Lemma~\ref{lem:lr_match}}

\noindent {\bf Proof}:

We just prove that is correct for $r_1$ and the proof for $r_2$ is nearly the same.

\begin{enumerate}
  \item If $r_1 \models filter(s, \alpha(r))$, then $filter(s, \alpha(r_1))=filter(s, \alpha(r))$ and $r_1 \models filter(s, \alpha(r_1))$. 

  Because $r_1 \models filter(s, \alpha(r))$, $\alpha(r_1)$ contains all the symbols in $filter(s, \alpha(r))$ (Definition of SOIRE matching). 
  Because $\alpha(r_1) \subseteq \alpha(r)$, $filter(s, \alpha(r_1))=filter(s, \alpha(r))$. 
  Therefore, $r_1 \models filter(s, \alpha(r_1))$.

  \item If $filter(s, \alpha(r_1))=filter(s, \alpha(r))$ and $r_1 \models filter(s, \alpha(r_1))$, then $r_1 \models filter(s, \alpha(r))$.

  Because $filter(s, \alpha(r_1))=filter(s, \alpha(r))$ and $r_1 \models filter(s, \alpha(r_1))$, $r_1 \models filter(s, \alpha(r))$.

\end{enumerate}

Lemma~\ref{lem:lr_match} has been proved.

\subsection{Proof of Theorem~\ref{thm:match}}

Theorem~\ref{thm:match_cond} shows that $r \models s$ iff $filter(s, \alpha(r))=s$ and $r \models filter(s, \alpha(r))$. 
In Algorithm~\ref{alg:SOIRETM}, Line~\ref{alg:SOIRETM_alpha} checks if $filter(s, \alpha(r))=s$. 
Therefore, we need to prove that $g_{1, |s|}^1=1$ iff $r \models filter(s, \alpha(r))$.
Algorithm~\ref{alg:SOIRETM} calculate $g_{i, j}^t$ for $s_{i, j}$ from short to long and for $r^t$ from bottom to top of the syntax tree. 
We apply mathematical induction to the length of the substring $|s_{i, j}|$ and structural induction to $r^t$. 

\noindent {\bf Proof}:

{\bf Mathematical induction base}: $|s_{i, j}|=l=0$, i.e. $s_{i, j}=\epsilon$ and $i=1, j=0$.

{\bf Structural induction}: Each case is corresponding to one statement in Line \ref{alg:SOIRETM_opr1}-\ref{alg:SOIRETM_opr7} in Algorithm~\ref{alg:SOIRETM}.

\begin{enumerate}
  \item $r^t=a \in \alpha(r)$.

  $r^t=a$ does not match $filter(s_{1, 0}, \{a\})=\epsilon$, so $g_{1, 0}^t=1[filter(s_{1, 0}, \{a\})=a]=0$ is correct.

  \item $r^t=(r^{t+1})^?$.
  
  $(r^{t+1})^? \models filter(s_{1, 0}, \alpha((r^{t+1})^?))=\epsilon$, so $g_{1, 0}^t=g_{1, 0}^{t+1}\ \vee\ 1[filter(s_{1, 0}, \alpha((r^{t+1})^?))=\epsilon]=1$ is correct.

  \item $r^t=(r^{t+1})^*$.
  
  $(r^{t+1})^* \models filter(s_{1, 0}, \alpha((r^{t+1})^*))=\epsilon$, so $g_{1, 0}^t=1[filter(s_{1, 0}, \alpha(r^t))=\epsilon] \vee g_{1, 0}^{t+1} \vee \vee_{k=i}^{j-1} (g_{i, k}^t\ \wedge\ g_{k+1, j}^{t+1})=1$ is correct.

  \item $r^t=(r^{t+1})^+$.
  
  Because $i=1>j-1=-1$, $\vee_{k=i}^{j-1} (g_{i, k}^t\ \wedge\ g_{k+1, j}^{t+1})=0$. 
  Therefore $g_{1, 0}^t=g_{1, 0}^{t+1} \vee \vee_{k=i}^{j-1} (g_{i, k}^t\ \wedge\ g_{k+1, j}^{t+1})=g_{1, 0}^{t+1}$. 
  Because $g_{1, 0}^t=1$ iff $(r^{t+1})^+ \models \epsilon$, and $(r^{t+1})^+ \models \epsilon$ iff $r^{t+1} \models \epsilon$, $g_{1, 0}^t=1$ iff $r^{t+1} \models \epsilon$.
  Because $r^{t+1} \models \epsilon$ iff $g_{1, 0}^{t+1}=1$ (Structural inductive hypothesis), $g_{1, 0}^t=1$ iff $g_{1, 0}^{t+1}=1$. 
  Therefore, $g_{1, 0}^t=g_{1, 0}^{t+1} \vee \vee_{k=i}^{j-1} (g_{i, k}^t\ \wedge\ g_{k+1, j}^{t+1})$ is correct.

  \item $r^t=r^{t+1} \cdot r^{\eta^t}$.
  
  Because $i=1>j-1=-1$, $\vee_{k=i}^{j-1} (1[filter(s_{i, k}, \alpha(r^{t+1} \cdot r^{\eta^t}))=filter(s_{i, k}, \alpha(r^{t+1}))]\ \wedge\ g_{i, k}^{t+1}\ \wedge\ 1[filter(s_{k+1, j}, \alpha(r^{t+1} \cdot r^{\eta^t}))=filter(s_{k+1, j}, \alpha(r^{\eta^t}))]\ \wedge\ g_{k+1, j}^{\eta^t})=0$. 
  In addition, $filter(s_{1, 0}, \alpha(r^{t+1}))=\epsilon, filter(s_{1, 0}, \alpha(r^{\eta^t}))=\epsilon$, i.e. $filter(s_{1, 0}, \alpha(r^{t+1} \cdot r^{\eta^t}))=filter(s_{1, 0}, \alpha(r^{t+1}))=filter(s_{1, 0}, \alpha(r^{\eta^t}))=\epsilon$. 
  Therefore, $g_{1, 0}^t=(1[filter(s_{1, 0}, \alpha(r^{t+1} \cdot r^{\eta^t}))=filter(s_{1, 0}, \alpha(r^{t+1}))]\ \wedge\ g_{1, 0}^{t+1}\ \wedge\ g_{1, 0}^{\eta^t})\ \vee\ (1[filter(s_{1, 0}, \alpha(r^{t+1} \cdot r^{\eta^t}))=filter(s_{1, 0}, \alpha(r^{\eta^t}))]\ \wedge\ g_{1, 0}^{\eta^t}\ \wedge\ g_{1, 0}^{t+1}) \vee \vee_{k=i}^{j-1} (1[filter(s_{i, k}, \alpha(r^{t+1} \cdot r^{\eta^t}))=filter(s_{i, k}, \alpha(r^{t+1}))]\ \wedge\ g_{i, k}^{t+1}\ \wedge\ 1[filter(s_{k+1, j}, \alpha(r^{t+1} \cdot r^{\eta^t}))=filter(s_{k+1, j}, \alpha(r^{\eta^t}))]\ \wedge\ g_{k+1, j}^{\eta^t})=g_{1, 0}^{t+1}\ \wedge\ g_{1, 0}^{\eta^t}$. 
  Because $g_{1, 0}^t=1$ iff $r^{t+1} \cdot r^{\eta^t} \models \epsilon$, and $r^{t+1} \cdot r^{\eta^t} \models \epsilon$ iff both $r^{t+1}, r^{\eta^t} \models \epsilon$, $g_{1, 0}^t=1$ iff both $r^{t+1}, r^{\eta^t} \models \epsilon$.
  Because $r^{t+1}$ (reps. $r^{\eta^t}$) matches $\epsilon$ iff $g_{1, 0}^{t+1}=1$ (reps. $g_{1, 0}^{\eta^t}=1$) (Structural inductive hypothesis), $g_{1, 0}^t=1$ iff $(g_{1, 0}^{t+1} \wedge g_{1, 0}^{\eta^t})=1$. 
  Therefore, $g_{1, 0}^t=(1[filter(s_{1, 0}, \alpha(r^{t+1} \cdot r^{\eta^t}))=filter(s_{1, 0}, \alpha(r^{t+1}))]\ \wedge\ g_{1, 0}^{t+1}\ \wedge\ g_{1, 0}^{\eta^t})\ \vee\ (1[filter(s_{1, 0}, \alpha(r^{t+1} \cdot r^{\eta^t}))=filter(s_{1, 0}, \alpha(r^{\eta^t}))]\ \wedge\ g_{1, 0}^{\eta^t}\ \wedge\ g_{1, 0}^{t+1}) \vee \vee_{k=i}^{j-1} (1[filter(s_{i, k}, \alpha(r^{t+1} \cdot r^{\eta^t}))=filter(s_{i, k}, \alpha(r^{t+1}))]\ \wedge\ g_{i, k}^{t+1}\ \wedge\ 1[filter(s_{k+1, j}, \alpha(r^{t+1} \cdot r^{\eta^t}))=filter(s_{k+1, j}, \alpha(r^{\eta^t}))]\ \wedge\ g_{k+1, j}^{\eta^t})$ is correct.

  \item $r^t=r^{t+1} \& r^{\eta^t}$.
  
  Because $g_{1, 0}^t=1$ iff $r^{t+1} \& r^{\eta^t} \models \epsilon$, and $r^{t+1} \& r^{\eta^t} \models \epsilon$ iff both $r^{t+1}, r^{\eta^t} \models \epsilon$, $g_{1, 0}^t=1$ iff both $r^{t+1}, r^{\eta^t} \models \epsilon$.
  Because $r^{t+1}$ (reps. $r^{\eta^t}$) matches $\epsilon$ iff $g_{1, 0}^{t+1}=1$ (reps. $g_{1, 0}^{\eta^t}=1$) (Structural inductive hypothesis), $g_{1, 0}^t=1$ iff $(g_{1, 0}^{t+1} \wedge g_{1, 0}^{\eta^t})=1$. 
  Therefore, $g_{1, 0}^t=g_{1, 0}^{t+1}\ \wedge\ g_{1, 0}^{\eta^t}$ is correct.

  \item $r^t=r^{t+1} | r^{\eta^t}$.
  
  Because $filter(s_{1, 0}, \alpha(r^{t+1}))=\epsilon, filter(s_{1, 0}, \alpha(r^{\eta^t}))=\epsilon$, i.e. $filter(s_{1, 0}, \alpha(r^{t+1} | r^{\eta^t}))=filter(s_{1, 0}, \alpha(r^{t+1}))=filter(s_{1, 0}, \alpha(r^{\eta^t}))=\epsilon$, $g_{1, 0}^t=(1[filter(s_{1, 0}, \alpha(r^{t+1} | r^{\eta^t}))=filter(s_{1, 0}, \alpha(r^{t+1}))]\ \wedge\ g_{1, 0}^{t+1})\ \vee\ (1[filter(s_{1, 0}, \alpha(r^{t+1} | r^{\eta^t}))=filter(s_{1, 0}, \alpha(r^{\eta^t}))]\ \wedge\ g_{1, 0}^{\eta^t})=g_{1, 0}^{t+1}\ \vee\ g_{1, 0}^{\eta^t}$. 
  Because $g_{1, 0}^t=1$ iff $r^{t+1} | r^{\eta^t} \models \epsilon$, and $r^{t+1} | r^{\eta^t} \models \epsilon$ iff $r^{t+1}$ or $r^{\eta^t} \models \epsilon$, $g_{1, 0}^t=1$ iff $r^{t+1}$ or $r^{\eta^t} \models \epsilon$. 
  Because $r^{t+1}$ (reps. $r^{\eta^t}$) matches $\epsilon$ iff $g_{1, 0}^{t+1}=1$ (reps. $g_{1, 0}^{\eta^t}=1$) (Structural inductive hypothesis), $g_{1, 0}^t=1$ iff $(g_{1, 0}^{t+1} \vee g_{1, 0}^{\eta^t})=1$. 
  Therefore, $g_{1, 0}^t=(1[filter(s_{1, 0}, \alpha(r^{t+1} | r^{\eta^t}))=filter(s_{1, 0}, \alpha(r^{t+1}))]\ \wedge\ g_{1, 0}^{t+1})\ \vee\ (1[filter(s_{1, 0}, \alpha(r^{t+1} | r^{\eta^t}))=filter(s_{1, 0}, \alpha(r^{\eta^t}))]\ \wedge\ g_{1, 0}^{\eta^t})$ is correct.

\end{enumerate}

{\bf Mathematical induction step}: $|s_{i, j}|=l>0$. Suppose that for all $s_{i', j'}$ ($|s_{i', j'}|=l'<l$) and $1 \leq t \leq |r|$, $g_{i', j'}^t=1$ iff $r^t \models s_{i', j'}$.

{\bf Structural induction}: Each case is corresponding to one statement in Line \ref{alg:SOIRETM_opr1}-\ref{alg:SOIRETM_opr7} in Algorithm~\ref{alg:SOIRETM}.

\begin{enumerate}
  \item $r^t=a \in \alpha(r)$.

  Because $g_{i, j}^t=1$ iff $r^t=a \in \alpha(r) \models filter(s_{i, j}, \{a\})$, i.e. $filter(s_{i, j}, \{a\})=a$, so $g_{i, j}^t=1$ iff $filter(s_{i, j}, \{a\})=a$. 
  Therefore, $g_{i, j}^t=1[filter(s_{i, j}, \{a\})=a]$ is correct.

  \item $r^t=(r^{t+1})^?$.
  
  $g_{i, j}^t=1$ iff $(r^{t+1})^? \models filter(s_{i, j}, \alpha((r^{t+1})^?))$, which is also equivalent to that $r^{t+1} \models filter(s_{i, j}, \alpha((r^{t+1})^?))$ or $filter(s_{i, j}, \alpha((r^{t+1})^?))=\epsilon$. 
  Because $\alpha((r^{t+1})^?)=\alpha(r^{t+1})$, $filter(s_{i, j}, \alpha((r^{t+1})^?))=filter(s_{i, j}, \alpha(r^{t+1}))$. 
  Therefore, $(r^{t+1})^? \models filter(s_{i, j}, \alpha((r^{t+1})^?))$ iff $r^{t+1} \models filter(s_{i, j}, \alpha(r^{t+1}))$ or $filter(s_{i, j}, \alpha((r^{t+1})^?))=\epsilon$. 
  $r^{t+1} \models filter(s_{i, j}, \alpha(r^{t+1}))$ iff $g_{i, j}^{t+1}=1$ (Structural inductive hypothesis), so $g_{i, j}^t=1$ iff $g_{i, j}^{t+1}=1$ or $1[filter(s_{i, j}, \alpha((r^{t+1})^?))=\epsilon]$. 
  Therefore, $g_{i, j}^t=g_{i, j}^{t+1} \vee 1[filter(s_{i, j}, \alpha((r^{t+1})^?))=\epsilon]$ is correct.

  \item $r^t=(r^{t+1})^*$.
  
  $g_{i, j}^t=1$ iff $(r^{t+1})^* \models filter(s_{i, j}, \alpha((r^{t+1})^*))$, which is also equivalent to that $filter(s_{i, j}, \alpha((r^{t+1})^*))=\epsilon$ or $r^{t+1} \models filter(s_{i, j}, \alpha((r^{t+1})^*))$, or $\exists i \leq k < j, (r^{t+1})^* \models filter(s_{i, k}, \alpha((r^{t+1})^*))$ and $r^{t+1} \models filter(s_{k+1, j}, \alpha((r^{t+1})^*))$. 
  Because $\alpha((r^{t+1})^*)=\alpha(r^{t+1})$, $filter(s_{i, j}, \alpha((r^{t+1})^*))=filter(s_{i, j}, \alpha(r^{t+1}))$ and $filter(s_{k+1, j}, \alpha((r^{t+1})^*))=filter(s_{k+1, j}, \alpha(r^{t+1}))$. 
  Therefore, $(r^{t+1})^* \models filter(s_{i, j}, \alpha((r^{t+1})^*))$ iff $filter(s_{i, j}, \alpha((r^{t+1})^*))=\epsilon$ or $r^{t+1} \models filter(s_{i, j}, \alpha(r^{t+1}))$, or $\exists i \leq k < j, (r^{t+1})^* \models filter(s_{i, k}, \alpha((r^{t+1})^*))$ and $r^{t+1} \models filter(s_{k+1, j}, \alpha(r^{t+1}))$. 
  $r^{t+1} \models filter(s_{i, j}, \alpha(r^{t+1}))$ iff $g_{i, j}^{t+1}=1$ (Structural inductive hypothesis). 
  Because $|s_{i, k}|=k-i+1<l$, $(r^{t+1})^* \models filter(s_{i, k}, \alpha((r^{t+1})^*))$ iff $g_{i, k}^t=1$ (Mathematical inductive hypothesis). 
  Because $|s_{k+1, j}|=j-(k+1)+1<l$, $r^{t+1} \models filter(s_{k+1, j}, \alpha(r^{t+1}))$ iff $g_{k+1, j}^{t+1}=1$ (Mathematical inductive hypothesis). 
  Therefore, $\exists i \leq k < j, (r^{t+1})^* \models filter(s_{i, k}, \alpha((r^{t+1})^*))$ and $r^{t+1} \models filter(s_{k+1, j}, \alpha(r^{t+1}))$ iff $\vee_{k=i}^{j-1} (g_{i, k}^t\ \wedge\ g_{k+1, j}^{t+1})=1$. 
  Therefore, $g_{i, j}^t=1$ iff $1[filter(s_{i, j}, \alpha(r^t))=\epsilon] \vee g_{i, j}^{t+1}=1 \vee \vee_{k=i}^{j-1} (g_{i, k}^t\ \wedge\ g_{k+1, j}^{t+1})=1$. 
  Therefore, $g_{i, j}^t=1[filter(s_{i, j}, \alpha(r^t))=\epsilon] \vee g_{i, j}^{t+1} \vee \vee_{k=i}^{j-1} (g_{i, k}^t\ \wedge\ g_{k+1, j}^{t+1})$ is correct.

  \item $r^t=(r^{t+1})^+$.
  
  $g_{i, j}^t=1$ iff $(r^{t+1})^+ \models filter(s_{i, j}, \alpha((r^{t+1})^+))$, which is also equivalent to that $r^{t+1} \models filter(s_{i, j}, \alpha((r^{t+1})^+))$, or $\exists i \leq k < j, (r^{t+1})^+ \models filter(s_{i, k}, \alpha((r^{t+1})^+))$ and $r^{t+1} \models filter(s_{k+1, j}, \alpha((r^{t+1})^+))$. 
  Because $\alpha((r^{t+1})^+)=\alpha(r^{t+1})$, $filter(s_{i, j}, \alpha((r^{t+1})^+))=filter(s_{i, j}, \alpha(r^{t+1}))$ and $filter(s_{k+1, j}, \alpha((r^{t+1})^+))=filter(s_{k+1, j}, \alpha(r^{t+1}))$. 
  Therefore, $(r^{t+1})^+ \models filter(s_{i, j}, \alpha((r^{t+1})^+))$ iff $r^{t+1} \models filter(s_{i, j}, \alpha(r^{t+1}))$, or $\exists i \leq k < j, (r^{t+1})^+ \models filter(s_{i, k}, \alpha((r^{t+1})^+))$ and $r^{t+1} \models filter(s_{k+1, j}, \alpha(r^{t+1}))$. 
  $r^{t+1} \models filter(s_{i, j}, \alpha(r^{t+1}))$ iff $g_{i, j}^{t+1}=1$ (Structural inductive hypothesis). 
  Because $|s_{i, k}|=k-i+1<l$, $(r^{t+1})^+ \models filter(s_{i, k}, \alpha((r^{t+1})^+))$ iff $g_{i, k}^t=1$ (Mathematical inductive hypothesis). 
  Because $|s_{k+1, j}|=j-(k+1)+1<l$, $r^{t+1} \models filter(s_{k+1, j}, \alpha(r^{t+1}))$ iff $g_{k+1, j}^{t+1}=1$ (Mathematical inductive hypothesis). 
  Therefore, $\exists i \leq k < j, (r^{t+1})^+ \models filter(s_{i, k}, \alpha((r^{t+1})^+))$ and $r^{t+1} \models filter(s_{k+1, j}, \alpha(r^{t+1}))$ iff $\vee_{k=i}^{j-1} (g_{i, k}^t\ \wedge\ g_{k+1, j}^{t+1})=1$. 
  Therefore, $g_{i, j}^t=1$ iff $(g_{i, j}^{t+1} \vee \vee_{k=i}^{j-1} (g_{i, k}^t\ \wedge\ g_{k+1, j}^{t+1}))=1$. 
  Therefore, $g_{i, j}^t=g_{i, j}^{t+1} \vee \vee_{k=i}^{j-1} (g_{i, k}^t\ \wedge\ g_{k+1, j}^{t+1})$ is correct.

  \item $r^t=r^{t+1} \cdot r^{\eta^t}$.
  
  $g_{i, j}^t=1$ iff $r^{t+1} \cdot r^{\eta^t} \models filter(s_{i, j}, \alpha(r^{t+1} \cdot r^{\eta^t}))$, which is also equivalent to that $r^{t+1} \models filter(s_{i, j}, \alpha(r^{t+1} \cdot r^{\eta^t}))$ and $r^{\eta^t} \models \epsilon$, or $r^{\eta^t} \models filter(s_{i, j}, \alpha(r^{t+1} \cdot r^{\eta^t}))$ and $r^{t+1} \models \epsilon$, or $\exists i \leq k < j, r^{t+1} \models filter(s_{i, k}, \alpha(r^{t+1} \cdot r^{\eta^t}))$ and $r^{\eta^t} \models filter(s_{k+1, j}, \alpha(r^{t+1} \cdot r^{\eta^t}))$. 
  $r^{t+1} \models filter(s_{i, j}, \alpha(r^{t+1} \cdot r^{\eta^t}))$ iff $filter(s_{i, j}, \alpha(r^{t+1}))=filter(s_{i, j}, \alpha(r^{t+1} \cdot r^{\eta^t}))$ and $r^{t+1} \models filter(s_{i, j}, \alpha(r^{t+1}))$ (Lemma~\ref{lem:lr_match}). 
  $r^{\eta^t} \models filter(s_{i, j}, \alpha(r^{t+1} \cdot r^{\eta^t}))$ iff $filter(s_{i, j}, \alpha(r^{\eta^t}))=filter(s_{i, j}, \alpha(r^{t+1} \cdot r^{\eta^t}))$ and $r^{\eta^t} \models filter(s_{i, j}, \alpha(r^{\eta^t}))$ (Lemma~\ref{lem:lr_match}). 
  Therefore, $r^{t+1} \cdot r^{\eta^t} \models filter(s_{i, j}, \alpha(r^{t+1} \cdot r^{\eta^t}))$ iff $filter(s_{i, j}, \alpha(r^{t+1}))=filter(s_{i, j}, \alpha(r^{t+1} \cdot r^{\eta^t}))$ and $r^{t+1} \models filter(s_{i, j}, \alpha(r^{t+1}))$ and $r^{\eta^t} \models \epsilon$, or $filter(s_{i, j}, \alpha(r^{\eta^t}))=filter(s_{i, j}, \alpha(r^{t+1} \cdot r^{\eta^t}))$ and $r^{\eta^t} \models filter(s_{i, j}, \alpha(r^{\eta^t}))$ and $r^{t+1} \models \epsilon$, or $\exists i \leq k < j,\ filter(s_{i, k}, \alpha(r^{t+1}))=filter(s_{i, k}, \alpha(r^{t+1} \cdot r^{\eta^t}))$ and $r^{t+1} \models filter(s_{i, k}, \alpha(r^{t+1}))$ and $filter(s_{k+1, j}, \alpha(r^{\eta^t}))=filter(s_{k+1, j}, \alpha(r^{t+1} \cdot r^{\eta^t}))$ and $r^{\eta^t} \models filter(s_{k+1, j}, \alpha(r^{\eta^t}))$. 
  $r^{t+1} \models filter(s_{i, j}, \alpha(r^{t+1}))$ iff $g_{i, j}^{t+1}=1$ (Structural inductive hypothesis). 
  $r^{\eta^t} \models filter(s_{i, j}, \alpha(r^{\eta^t}))$ iff $g_{i, j}^{\eta^t}=1$ (Structural inductive hypothesis). 
  $r^{t+1}$ (reps. $r^{\eta^t}$) matches $\epsilon$ iff $g_{1, 0}^{t+1}=1$ (reps. $g_{1, 0}^{\eta^t}=1$) (Mathematical inductive hypothesis). 
  Because $|s_{i, k}|=k-i+1<l$, $r^{t+1} \models filter(s_{i, k}, \alpha(r^{t+1}))$ iff $g_{i, k}^{t+1}=1$ (Mathematical inductive hypothesis). 
  Because $|s_{k+1, j}|=j-(k+1)+1<l$, $r^{\eta^t} \models filter(s_{k+1, j}, \alpha(r^{\eta^t}))$ iff $g_{k+1, j}^{\eta^t}=1$ (Mathematical inductive hypothesis). 
  Therefore, $\exists i \leq k < j,\ filter(s_{i, k}, \alpha(r^{t+1}))=filter(s_{i, k}, \alpha(r^{t+1} \cdot r^{\eta^t}))$ and $r^{t+1} \models filter(s_{i, k}, \alpha(r^{t+1}))$ and $filter(s_{k+1, j}, \alpha(r^{\eta^t}))=filter(s_{k+1, j}, \alpha(r^{t+1} \cdot r^{\eta^t}))$ and $r^{\eta^t} \models filter(s_{k+1, j}, \alpha(r^{\eta^t}))$ iff $\vee_{k=i}^{j-1} (1[filter(s_{i, k}, \alpha(r^{t+1} \cdot r^{\eta^t}))=filter(s_{i, k}, \alpha(r^{t+1}))]\ \wedge\ g_{i, k}^{t+1}\ \wedge\ 1[filter(s_{k+1, j}, \alpha(r^{t+1} \cdot r^{\eta^t}))=filter(s_{k+1, j}, \alpha(r^{\eta^t}))]\ \wedge\ g_{k+1, j}^{\eta^t})=1$. 
  Therefore, $g_{i, j}^t=1$ iff $(1[filter(s_{i, j}, \alpha(r^{t+1} \cdot r^{\eta^t}))=filter(s_{i, j}, \alpha(r^{t+1}))]\ \wedge\ g_{i, j}^{t+1}\ \wedge\ g_{1, 0}^{\eta^t})\ \vee\ (1[filter(s_{i, j}, \alpha(r^{t+1} \cdot r^{\eta^t}))=filter(s_{i, j}, \alpha(r^{\eta^t}))]\ \wedge\ g_{i, j}^{\eta^t}\ \wedge\ g_{1, 0}^{t+1}) \vee \vee_{k=i}^{j-1} (1[filter(s_{i, k}, \alpha(r^{t+1} \cdot r^{\eta^t}))=filter(s_{i, k}, \alpha(r^{t+1}))]\ \wedge\ g_{i, k}^{t+1}\ \wedge\ 1[filter(s_{k+1, j}, \alpha(r^{t+1} \cdot r^{\eta^t}))=filter(s_{k+1, j}, \alpha(r^{\eta^t}))]\ \wedge\ g_{k+1, j}^{\eta^t})=1$. 
  Therefore, $g_{i, j}^t=(1[filter(s_{i, j}, \alpha(r^{t+1} \cdot r^{\eta^t}))=filter(s_{i, j}, \alpha(r^{t+1}))]\ \wedge\ g_{i, j}^{t+1}\ \wedge\ g_{1, 0}^{\eta^t})\ \vee\ (1[filter(s_{i, j}, \alpha(r^{t+1} \cdot r^{\eta^t}))=filter(s_{i, j}, \alpha(r^{\eta^t}))]\ \wedge\ g_{i, j}^{\eta^t}\ \wedge\ g_{1, 0}^{t+1}) \vee \vee_{k=i}^{j-1} (1[filter(s_{i, k}, \alpha(r^{t+1} \cdot r^{\eta^t}))=filter(s_{i, k}, \alpha(r^{t+1}))]\ \wedge\ g_{i, k}^{t+1}\ \wedge\ 1[filter(s_{k+1, j}, \alpha(r^{t+1} \cdot r^{\eta^t}))=filter(s_{k+1, j}, \alpha(r^{\eta^t}))]\ \wedge\ g_{k+1, j}^{\eta^t})$ is correct.

  \item $r^t=r^{t+1} \& r^{\eta^t}$.
  
  $g_{i, j}^t=1$ iff $r^{t+1} \& r^{\eta^t} \models filter(s_{i, j}, \alpha(r^{t+1} \& r^{\eta^t}))$. 
  Because each symbol only occurs once in an SOIRE and $r^t, r^{t+1}, r^{\eta^t}$ are SOIREs, $\alpha(r^t)=\alpha(r^{t+1}) \cup \alpha(r^{\eta^t}), \alpha(r^{t+1}) \cap \alpha(r^{\eta^t})=\emptyset$. 
  Therefore, each symbol in $filter(s_{i, j}, \alpha(r^{t+1} \& r^{\eta^t}))$ is either in $\alpha(r^{t+1})$ or $\alpha(r^{\eta^t})$. 
  $r^{t+1} \& r^{\eta^t} \models filter(s_{i, j}, \alpha(r^{t+1} \& r^{\eta^t}))$ iff $filter(s_{i, j}, \alpha(r^{t+1} \& r^{\eta^t}))$ can be devided into two subsequences and $r^{t+1}$ matches one subsequence and $r^{\eta^t}$ matches the other. 
  Because each symbol in $filter(s_{i, j}, \alpha(r^{t+1} \& r^{\eta^t}))$ is either in $\alpha(r^{t+1})$ or $\alpha(r^{\eta^t})$, one subsequence consists of all the symbols in $\alpha(r^{t+1})$, i.e. $filter(s_{i, j}, \alpha(r^{t+1}))$, which is matched by $r^{t+1}$, and the other subsequence consists of all the symbols in $\alpha(r^{\eta^t})$, i.e. $filter(s_{i, j}, \alpha(r^{\eta^t}))$, which is matched by $r^{\eta^t}$. 
  Therefore, $r^{t+1} \& r^{\eta^t} \models filter(s_{i, j}, \alpha(r^{t+1} \& r^{\eta^t}))$ iff $r^{t+1} \models filter(s_{i, j}, \alpha(r^{t+1}))$ and $r^{\eta^t} \models filter(s_{i, j}, \alpha(r^{\eta^t}))$. 
  $r^{t+1} \models filter(s_{i, j}, \alpha(r^{t+1}))$ iff $g_{i, j}^{t+1}=1$ (Structural inductive hypothesis).
  $r^{\eta^t} \models filter(s_{i, j}, \alpha(r^{\eta^t}))$ iff $g_{i, j}^{\eta^t}=1$ (Structural inductive hypothesis).
  Therefore $g_{i, j}^t=1$ iff $(g_{i, j}^{t+1}\ \wedge\ g_{i, j}^{\eta^t})=1$. 
  Therefore $g_{i, j}^t=g_{i, j}^{t+1}\ \wedge\ g_{i, j}^{\eta^t}$ is correct. 

  \item $r^t=r^{t+1} | r^{\eta^t}$.
  
  $g_{i, j}^t=1$ iff $r^{t+1} | r^{\eta^t} \models filter(s_{i, j}, \alpha(r^{t+1} | r^{\eta^t}))$, which is equivalent to that $r^{t+1} \models filter(s_{i, j}, \alpha(r^{t+1} | r^{\eta^t}))$ or $r^{\eta^t} \models filter(s_{i, j}, \alpha(r^{t+1} | r^{\eta^t}))$. 
  $r^{t+1} \models filter(s_{i, j}, \alpha(r^{t+1} | r^{\eta^t}))$ iff $filter(s_{i, j}, \alpha(r^{t+1}))=filter(s_{i, j}, \alpha(r^{t+1} | r^{\eta^t}))$ and $r^{t+1} \models filter(s_{i, j}, \alpha(r^{t+1}))$ (Lemma~\ref{lem:lr_match}). 
  $r^{\eta^t} \models filter(s_{i, j}, \alpha(r^{t+1} | r^{\eta^t}))$ iff $filter(s_{i, j}, \alpha(r^{\eta^t}))=filter(s_{i, j}, \alpha(r^{t+1} | r^{\eta^t}))$ and $r^{\eta^t} \models filter(s_{i, j}, \alpha(r^{\eta^t}))$ (Lemma~\ref{lem:lr_match}). 
  Therefore, $r^{t+1} | r^{\eta^t} \models filter(s_{i, j}, \alpha(r^{t+1} | r^{\eta^t}))$ iff $filter(s_{i, j}, \alpha(r^{t+1}))=filter(s_{i, j}, \alpha(r^{t+1} | r^{\eta^t}))$ and $r^{t+1} \models filter(s_{i, j}, \alpha(r^{t+1}))$, or $filter(s_{i, j}, \alpha(r^{\eta^t}))=filter(s_{i, j}, \alpha(r^{t+1} | r^{\eta^t}))$ and $r^{\eta^t} \models filter(s_{i, j}, \alpha(r^{\eta^t}))$. 
  $r^{t+1} \models filter(s_{i, j}, \alpha(r^{t+1}))$ iff $g_{i, j}^{t+1}=1$ (Structural inductive hypothesis).
  $r^{\eta^t} \models filter(s_{i, j}, \alpha(r^{\eta^t}))$ iff $g_{i, j}^{\eta^t}=1$ (Structural inductive hypothesis). 
  Therefore, $g_{i, j}^t=1$ iff $(1[filter(s_{i, j}, \alpha(r^{t+1} | r^{\eta^t}))=filter(s_{i, j}, \alpha(r^{t+1}))]\ \wedge\ g_{i, j}^{t+1})\ \vee\ (1[filter(s_{i, j}, \alpha(r^{t+1} | r^{\eta^t}))=filter(s_{i, j}, \alpha(r^{\eta^t}))]\ \wedge\ g_{i, j}^{\eta^t})=1$. 
  Therefore, $g_{i, j}^t=(1[filter(s_{i, j}, \alpha(r^{t+1} | r^{\eta^t}))=filter(s_{i, j}, \alpha(r^{t+1}))]\ \wedge\ g_{i, j}^{t+1})\ \vee\ (1[filter(s_{i, j}, \alpha(r^{t+1} | r^{\eta^t}))=filter(s_{i, j}, \alpha(r^{\eta^t}))]\ \wedge\ g_{i, j}^{\eta^t})$ is correct.

\end{enumerate}

Theorem~\ref{thm:match} has been proved.

\subsection{Proof of Proposition~\ref{thm:img_parser}}

\noindent {\bf Proof}:

Algorithm~\ref{alg:prefix2middle} shows the properties of the prefix notation of an SOIRE. 
Consider the number of elements of the stack, denoted as $|\Gamma|$. 
If $o^i \in \Sigma$, $|\Gamma|$ adds $1$. 
If $o^i \in \{?, *, +\}$, $|\Gamma| \geq 1$ and after that $|\Gamma|$ stays unchanged.
If $o^i \in \{\cdot, \&, |\}$, $|\Gamma| \geq 2$ and after that $|\Gamma|$ minus $1$. 
Finally, $|\Gamma|=1$.
Therefore, 

\begin{align}\label{equ:prefix_property}
  &\forall 1 \leq i \leq |r_p|, (\sum_{j=i}^{|r_p|} 1[o^j \in \Sigma]-1[o^j \in \{\cdot, \&, |\}]) \geq 1 \nonumber\\
  &\wedge (\sum_{i=1}^{|r_p|} 1[o^i \in \Sigma]-1[o^i \in \{\cdot, \&, |\}]) = 1
\end{align}

We shows that $\encre(\theta)$ satisfies Equation~\ref{equ:prefix_property}, given the faithful encoding $\theta=(w, u)$. 
Let $last$ be the biggest number that $w^{last}_{\text{none}}=0$ ($|r_p|=last$) and $T$ be the bounded size. 
We first use mathematical induction to prove $\forall 1 \leq i \leq |r_p|, (\sum_{j=i}^{|r_p|} 1[o^j \in \Sigma]-1[o^j \in \{\cdot, \&, |\}]) \geq 1$. 

{\bf Mathematical induction base}: $i=last$.

If $i=last=T$, then $\sum_{a \in \Sigma} w^i_a=1$ (the last vertex can only select a symbol), i.e. $o^i \in \Sigma$. 
Therefore, $(\sum_{j=i}^{last} 1[o^j \in \Sigma]-1[o^j \in \{\cdot, \&, |\}]) \geq 1$ is correct. 
If $i=last<T$, because $w^{i+1}_{\text{none}}=1$, $\forall a \in \{?, *, +, \cdot, \&, |\}, w^i_{a}=0$ (Definition~\ref{def:faithful} Condition~\ref{def:faithful_5}). 
Therefore, $\sum_{a \in \Sigma} w^i_a=1$ (Definition~\ref{def:faithful} Condition~\ref{def:faithful_1}), i.e. $o^i \in \Sigma$. 
Therefore, $(\sum_{j=i}^{last} 1[o^j \in \Sigma]-1[o^j \in \{\cdot, \&, |\}]) \geq 1$ is correct. 

{\bf Mathematical induction step}: $1 \leq i < last$. Suppose for $i+1$, $(\sum_{j=i+1}^{last} 1[o^j \in \Sigma]-1[o^j \in \{\cdot, \&, |\}]) \geq 1$. 

$(\sum_{j=i+1}^{last} 1[o^j \in \Sigma]-1[o^j \in \{\cdot, \&, |\}]) \geq 1$ (Mathematical inductive hypothesis). 
If $o^i \in \Sigma \cup \{?, *, +\}$, $(\sum_{j=i}^{last} 1[o^j \in \Sigma]-1[o^j \in \{\cdot, \&, |\}]) \geq 1$ is correct. 
If $o^i \in \{\cdot, \&, |\}$, consider two cases:
\begin{enumerate}
  \item $(\sum_{j=i+1}^{last} 1[o^j \in \Sigma]-1[o^j \in \{\cdot, \&, |\}]) > 1$
  
  $(\sum_{j=i+1}^{last} 1[o^j \in \Sigma]-1[o^j \in \{\cdot, \&, |\}])+1[o^i \in \Sigma]-1[o^i \in \{\cdot, \&, |\}]> 1-1$. 
  Therefore, $(\sum_{j=i}^{last} 1[o^j \in \Sigma]-1[o^j \in \{\cdot, \&, |\}]) > 1$ is correct.
  
  \item $(\sum_{j=i+1}^{last} 1[o^j \in \Sigma]-1[o^j \in \{\cdot, \&, |\}]) = 1$
  
  Let $B=\{ j | i \leq j \leq last \wedge o^j \in \{\cdot, \&, |\}\}$. 
  For all $j \in B$, $\exists j+2 \leq k \leq last, u^j_{k}=1$, denoted as $u^j=k$ (Definition~\ref{def:faithful} Condition~\ref{def:faithful_3}). 
  For any $j, k \in B$, if $j \neq k$, $u^j \neq u^k$ (Definition~\ref{def:faithful} Condition~\ref{def:faithful_2} and \ref{def:faithful_5}). 
  For all $j \in B$, $o^{u^j-1} \in \Sigma$ (Definition~\ref{def:faithful} Condition~\ref{def:faithful_5} and \ref{def:faithful_1}). 
  Because $o^i \in \{\cdot, \&, |\}$, $(\sum_{j=i}^{last} 1[o^j \in \Sigma]-1[o^j \in \{\cdot, \&, |\}]) = 0$. 
  Therefore, $\{u^j-1 | j \in B\}$ is the set of all $o \in \Sigma$.
  Therefore, $o^{last} \notin \Sigma$, which is a contradiction.
  Therefore, this case does not exist.

\end{enumerate}

Therefore, $\encre(\theta)$ satisfies $\forall 1 \leq i \leq |r_p|, (\sum_{j=i}^{|r_p|} 1[o^j \in \Sigma]-1[o^j \in \{\cdot, \&, |\}]) \geq 1$.

Let $B=\{ i | 1 \leq i \leq last \wedge o^i \in \{\cdot, \&, |\}\}$. 
For all $i \in B$, $\exists i+2 \leq j \leq last, u^i_{j}=1$, denoted as $u^i=j$ (Definition~\ref{def:faithful} Condition~\ref{def:faithful_3}). 
For any $i, j \in B$, if $i \neq j$, $u^i \neq u^j$ (Definition~\ref{def:faithful} Condition~\ref{def:faithful_2} and \ref{def:faithful_5}). 
For all $i \in B$, $o^{u^i-1} \in \Sigma$ (Definition~\ref{def:faithful} Condition~\ref{def:faithful_5} and \ref{def:faithful_1}). 
In addition, $last \notin \{u^i-1 | i \in B\}$ and $o^{last} \in \Sigma$. 
Therefore, $\encre(\theta)$ satisfies $(\sum_{i=1}^{|r_p|} 1[o^i \in \Sigma]-1[o^i \in \{\cdot, \&, |\}]) = 1$.

Therefore, $\encre(\theta)$ satisfies Equation~\ref{equ:prefix_property} and $\encre(\theta)$ returns the prefix notation of an SOIRE.

Proposition~\ref{thm:img_parser} has been proved.

\subsection{Proof of Proposition~\ref{thm:surjective}}

\noindent {\bf Proof}:

We construct the faithful encoding $\theta=(w, u)$, given $T \in \mathbb{Z}^+$, and an SOIRE $r (|r| \leq T)$ written in prefix notation. 
Construct the syntax tree of $r$, and $o^t (1 \leq t \leq |r|)$ denotes the symbol or operator represented by vectex $t$ and vertex $v^t$ is the right son of vertex $t$. 
Let $w=\mathbf{0}$ and $u=\mathbf{0}$ initially. 
$\forall |r| < t \leq T, w^{t}_{\text{none}} \gets 1$. 
$\forall 1 \leq t \leq |r|, w^{t}_{o^t} \gets 1$. 
$\forall 1 \leq t \leq |r| \wedge o^t \in \{\cdot, \&, |\}, u^t_{v^t} \gets 1$. 
Because $\theta=(w, u)$ is constructed based on the syntax tree of $r$, $\theta=(w, u)$ are faithful encoding and $\encre(\theta)=\preorder(r)$.

Proposition~\ref{thm:surjective} has been proved.

\subsection{Proof of Proposition~\ref{thm:injection}}

\noindent {\bf Proof}:

We apply proof by contradiction to prove this theorem. 
Given the bouned size of SOIREs $T \in \mathbb{Z}^+$, for any two different faithful encodings $\theta_1, \theta_2 (\theta_1 \neq \theta_2)$, $\encre(\theta_1) = \encre(\theta_2)$. 
Let $\theta_1=(w^{(1)}, u^{(1)}), \theta_2=(w^{(2)}, u^{(2)})$. 
Because $\encre(\theta_1) = \encre(\theta_2)$, $w^{(1)}=w^{(2)}$ (Algorithm~\ref{alg:faithful_interpret} and Definition~\ref{def:faithful} Condition~\ref{def:faithful_4}). 
Let $p$ be the biggest number that $u^{(1)p} \neq u^{(2)p}$. 
Because $w^{(1)}=w^{(2)}$, $u^{(1)p} = u^{(2)p} = \mathbf{0}$ or $\exists p+2 \leq t_1, t_2 \leq T, u^{(1)p}_{t_1}=1 \wedge u^{(2)p}_{t_2}=1$. 
In addition, $u^{(1)p} \neq u^{(2)p}$. Therefore, $\exists p+2 \leq t_1, t_2 \leq T, u^{(1)p}_{t_1}=1 \wedge u^{(2)p}_{t_2}=1$. 
Without loss of generality, let $t_1<t_2$. 
Because $u^{(1)p}_{t_1}=1$, $\forall a \in \{?, *, +, \cdot, \&, |\}, w^{(1)t_1-1}_{a}=0$ (Definition~\ref{def:faithful} Condition~\ref{def:faithful_5}). 
Because $w^{(1)}=w^{(2)}$, $\forall a \in \{?, *, +, \cdot, \&, |\}, w^{(2)t_1-1}_{a}=0$ (Definition~\ref{def:faithful} Condition~\ref{def:faithful_5}). 
Therefore, $\exists 1 \leq p_2 \leq t_1-2, u^{(2)p_2}_{t_1}=1$ (Definition~\ref{def:faithful} Condition~\ref{def:faithful_5}). 
Because $p$ be the biggest number that $u^{(1)p} \neq u^{(2)p}$, $1 \leq p_2 < p$. 
For $\theta_2$, because vertex $p_2$ has a right son, vertex $t_1$, then for all vertex $p'(p_2<p'<t_1)$, if vertex $p'$ has a right son, vertex $t'$, then $t'<t_1$ (Definition~\ref{def:faithful} Condition~\ref{def:faithful_6}). 
However, $p_2<p<t_1$ and vertex $p$ has a right son, vertex $t_2$, and $t_2>t_1$, which is a contradiction. 
Therefore, there does not exist two different faithful encoding $\theta_1, \theta_2 (\theta_1 \neq \theta_2)$, where the size of $\theta_1$ and $\theta_2$ is $T$, $\encre(\theta_1) = \encre(\theta_2)$. 

Proposition~\ref{thm:injection} has been proved.

\subsection{Proof of Theorem~\ref{thm:onetoone}}

\noindent {\bf Proof}:

Theorem~\ref{thm:onetoone} can be proved by combining Proposition~\ref{thm:img_parser}, Proposition~\ref{thm:injection} and Proposition~\ref{thm:surjective}. 
Therefore, Faithful encoding is one-to-one corresponding to the SOIRE in prefix notation for a certain size of the parameters.

\subsection{Proof of Proposition~\ref{thm:T_bound}}

\noindent {\bf Proof}:

Because $(r^?)^?=r^?, (r^?)^*=r^*, (r^?)^+=r^*, (r^*)^*=r^*, (r^*)^?=r^*, (r^*)^+=r^*, (r^+)^+=r^+, (r^+)^?=r^*, (r^+)^*=r^*$, the consecutive unary operators can be replaced by one. 
For an SOIRE $r$ over $\Sigma$, we use $r'$ ($r' \equiv r$) to denote the SOIRE without consecutive unary operators.
For the syntax tree of $r'$, there are no consecutive unary operators in $r'$, so only the parent of a vertex representing a symbol or a binary operator can represent a unary operator.
The Equation~\ref{equ:prefix_property} shows that the number of the binary operators is $|\alpha(r')|-1$ for $r'$.
Therefore, the size of the syntax tree of $r'$ is $(|\alpha(r')|+|\alpha(r')|-1)*2=4|\alpha(r')|-2 \leq 4|\Sigma|-2$.
Proposition~\ref{thm:surjective} shows that given $T=4|\Sigma|-2$, there exists a faithful encoding $\theta$ that $\encre(\theta)=\preorder(r')$, because $r' \leq T$. 
Therefore, given $T=4|\Sigma|-2$, for each SOIRE $r$ over $\Sigma$, there exists a faithful encoding $\theta$ that $\encre(\theta)=\preorder(r')$ and $\{s | r' \models s\}=\{s | r \models s\}$.

Proposition~\ref{thm:T_bound} has been proved.

\section{Appendix C}
\label{app:experiment}

\subsection{Regularization}

To make the set of parameters more faithful, we can add one regularization for each condition when training the neural network. 
We use Equation~\ref{equ:reg_loss_1}-\ref{equ:reg_loss_7} as regularization for each condition. 
$\text{Onehotloss}(x)=\text{Mean}_i (1-x_i)x_i+ (1-\sum_i x_i)^2$ for a vector $x$. 
$\text{ReLU}(x)=\max(x, 0)$ for a scalar $x$.

\begin{align}
  &\text{Mean}_{t=1}^{T}(\text{Onehotloss}(w^t)) \label{equ:reg_loss_1} \\
  &\text{Mean}_{t=1}^{T}(\text{Onehotloss}(u^t)) \\
  &\text{Mean}_{t=1}^{T}(\text{Onehotloss}([u^t_{t+2}, u^t_{t+3}, \dots, u^t_{T}, \nonumber \\
  &w^t_{a \in \Sigma}, w^t_{?}, w^t_{*}, w^t_{+}, w^t_{\text{none}}])) \\
  &\text{Mean}_{t=1}^{T-1}(\text{ReLU}(w^t_{\text{none}}-w^{t+1}_{\text{none}})) \\
  &\text{Mean}_{t=2}^{T}(\text{Onehotloss}([w^{t-1}_{?}, w^{t-1}_{+}, w^{t-1}_{*}, w^{t-1}_{\cdot}, \nonumber \\
  &w^{t-1}_{\&}, w^{t-1}_{|}, u^1_t, u^2_t, \dots, u^{t-2}_t, w^t_{\text{none}}]))\\
  &\text{Mean}_{t=3}^{T}(\text{Mean}_{p=1}^{t-2}(\text{ReLU}((t-1-p)u^p_t+ \nonumber \\
  &\sum_{p'=p+1}^{t-1}\sum_{t'=t+1}^{T}u^{p'}_{t'} - (t-1-p)))) \\
  &\text{Mean}_{a \in \Sigma}(\text{ReLU}(\sum_{t=1}^T w^t_a-1)) \label{equ:reg_loss_7}
\end{align}

\subsection{Results on noise-free data}

The SOIREs learnt by \SOIREDL on each dataset with noise-free data are show in Table~\ref{tab:result}.

\begin{table}[htbp]
    \centering
    \scalebox{0.80}{
        \begin{tabular}{|c|c|c|c|c|}
        \hline
        Dataset & Ground-truth SOIREs                                 & SOIREDL                                            \\
        \hline
        1  & $( ( a|b )c^{*} )^{+}d$            & $( ( a|b )c^{*} )^{+}d$       \\
        2  & $(a|b|c|d|e^?)^*$                  & $(b^*e^*a^*c^*d^*)^*$         \\
        3  & $a^{+} | ( b|c )^{*} | d^{+}$      & $( b^{+}|c )^{*} | a^{*} | d^{*}$        \\
        4  & $(ab^*)^+|c^+$                     & $(a^+b^*)^*$                \\
        5  & $ab(c|d|e|f|g)^*$                  & $((c|g^*|a\&b|d^*)e^*)^*$        \\
        6  & $a(b|c|d|e)^+f^*$                  & $(b^*\&a\&d^*\&e^*\&c^*)f^*$            \\
        7  & $a^*|(b|c|d|e)^*$                  & $(c^*d^*e^*)^*\&b^*$                  \\
        8  & $(a|b|c|d|e)^+\&f^*\&g^*$          & $(f^*g^*d^*a^*e^*c^*b^*)^+$             \\
        9  & $a^+|b|(c|d)^+$                    & $(c^+|d^+)^+$                       \\
        10 & $(a|b)^+c$                         & $((a^*\&b^*)^+c)^?$                  \\
        11 & $a^?(b|c|d)^*e$             	    & $(((b^*|d^*)a^*c^*)^+e)^?$             \\
        12 & $(a|b|c|d^?)^*$                    & $(a^*b^*d^*c^*)^*$                   \\
        13 & $a^{?} \& b^{*} \& c^{?}$          & $a^{?} \& c^{?} \& b^{*}$            \\
        14 & $(a(bc^?)^*)^*d^?e^?$              & $(a(b^+\&c)^*d^*e^?)^*$              \\
        15 & $a^?\&b^*\&c^*\&d^?$	            & $c^*\&a^?\&d\&b^*$              \\
        16 & $a^*\&b^?\&c^*$ 	                & $a^*\&c^*\&b$              \\
        17 & $a^+\&(b|c|d|e)^+$	                & $(c^*|e^*\&(a^*|b^+)|d^*)^*$       \\
        18 & $a^?(b|c)^*(d|e)$                  & $a^*(b|c)^*d^?e^?$               \\
        19 & $a(b|c|d)^*e^*$                    & $a((c|d)^?|b)^*e^*$                \\
        20 & $(a^+|b^?|c^?|(d|e)^*)fgh^*$       & $(d^*\&f\&g^*\&e^*)^?h^*$    \\
        21 & $(a^*b)^+$	                        & $(a^*b)^*$                 \\
        22 & $(a|b)^*c^*d^*$                    & $(a^*\&b^*)c^*d^*$            \\
        23 & $((ab^*)^+|c^+)d$                  & $(ab^*d^?)^+$               \\
        24 & $(a|b|c)^+|(de^*f^*)^+$            & $((c^*a^*)^*b^*)^+$          \\
        25 & $a|(b^?c)^+$                       & $(b^?c^+)^+$              \\
        26 & $a^*(b|c|d)^*$	                    & $((b^*|a^*d)c^*)^*$        \\
        27 & $a^?(b^+\&c^*\&d^*\&e^*\&f^*)$     & $((c^*\&d^*\&e^*)f^*b^*)^*$       \\
        28 & $(a^?b)^+$                         & $(a^?b)^*$           \\
        29 & $(a^+|b^+|(c|d)^*)e$               & $(d^*c^*)^*e$           \\
        30 & $a^?(b^?c^+d)^+$                   & $((a|b)^*cd^?)^*$        \\

        \hline
    \end{tabular}}
        \caption{The ground-truth SOIREs and the SOIREs learnt by SOIREDL on each dataset with noise-free data.}
        \label{tab:result}
    \end{table}

\subsection{Hyperparameters}
All hyperparameters are tuned on the first noise-free dataset.

\noindent{\bf Hyperparameters of \RERNN}

Max-state and $\gamma$ are two important hyperparameters of \RERNN. 
We use grid search to choose max-state from $\{$$20$, $40$, $60$, $80$, $100$, $120$, $140$, $160$, $180$, $200$$\}$ and choose $\gamma$ from $\{$$0$, $0.02$, $0.04$, $0.06$, $0.08$, $0.1$, $0.12$, $0.14$, $0.16$, $0.18$, $0.2$$\}$.

\begin{figure}[htbp]
  \includegraphics[width=0.43\textwidth]{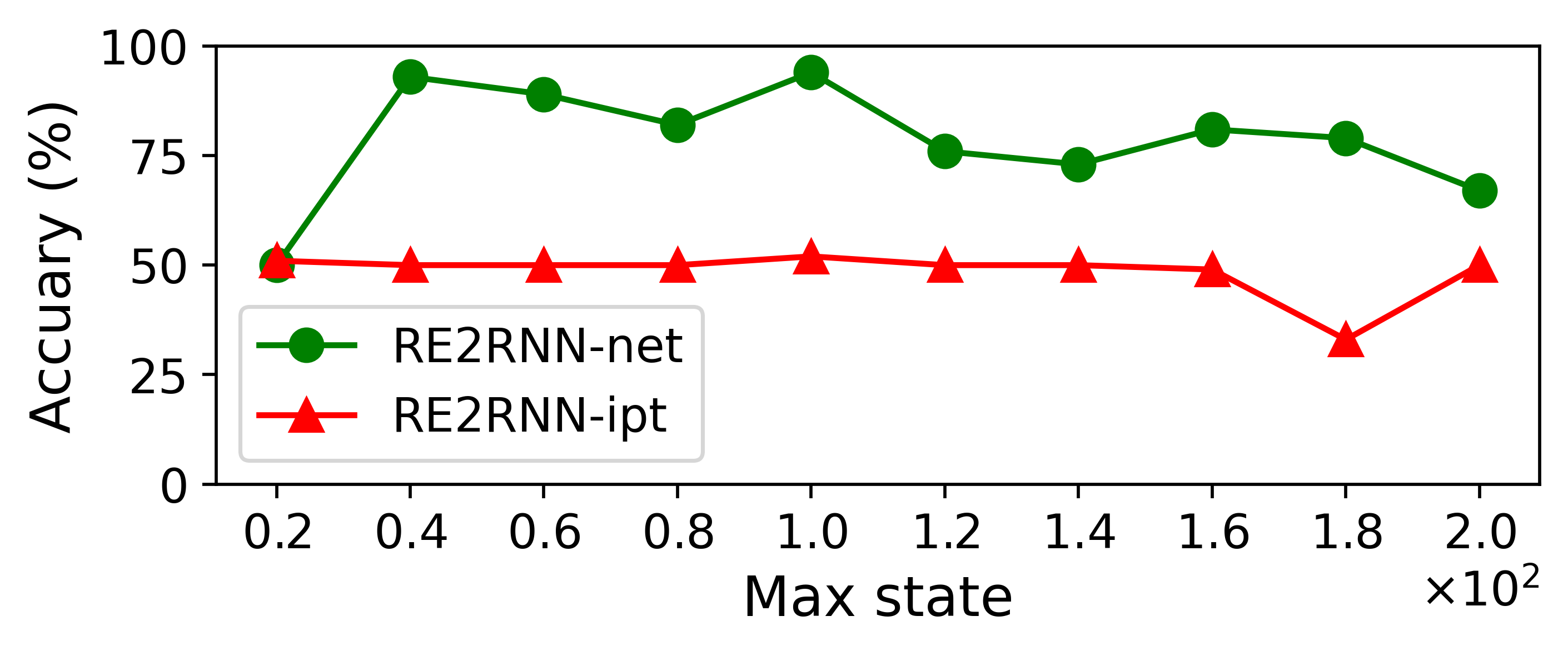}
  \caption{Accuracy(\%) of \RERNN on the first dataset with different number of states max-state.}
  \label{fig:max_state}
\end{figure}

Max-state is the number of state in the automaton and Figure~\ref{fig:max_state} shows that max-state should be set as $100$ to get the best performance. 

\begin{figure}[htbp]
  \includegraphics[width=0.43\textwidth]{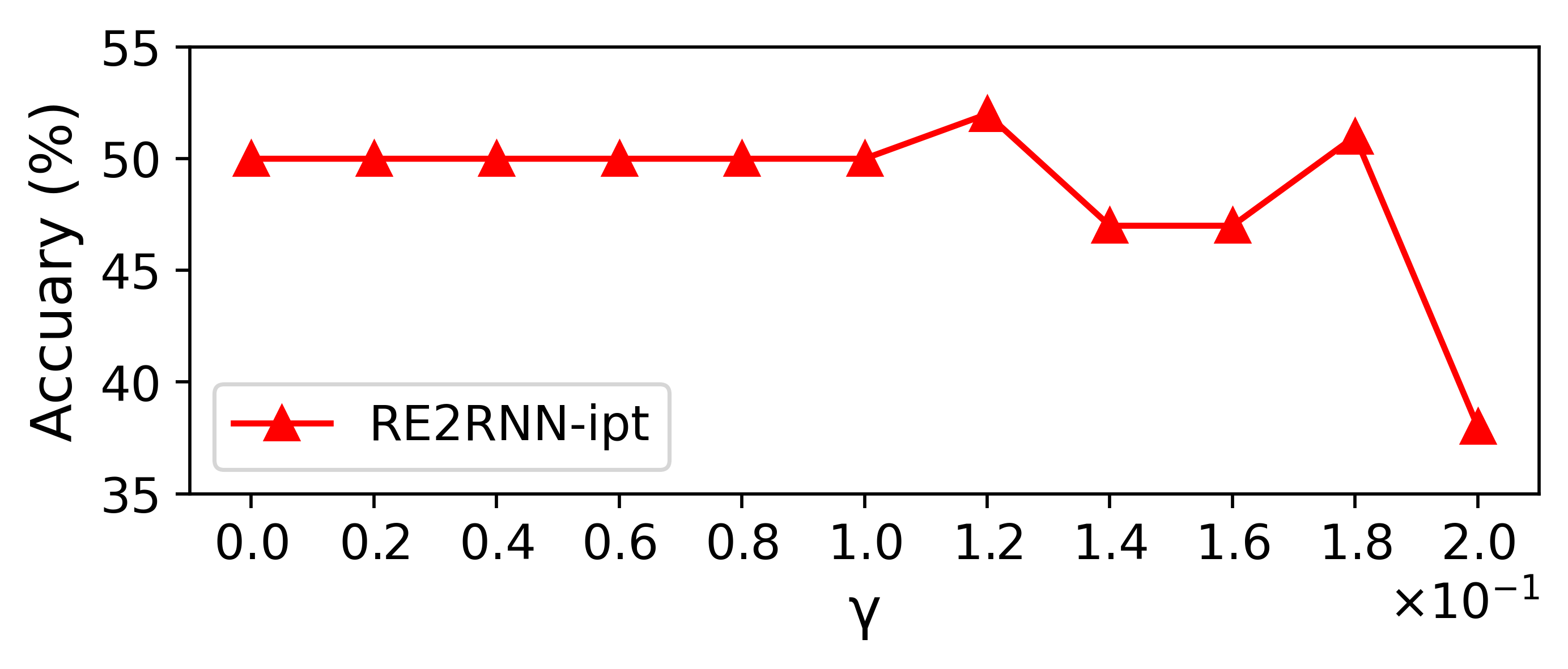}
  \caption{Accuracy(\%) of \RERNN on the first dataset with different thresholds in the interpretation $\gamma$.}
  \label{fig:gamma}
\end{figure}

$\gamma$ is the threshold in the interpretation. 
The values above $\gamma$ are set as $1$s and the values below $\gamma$ are set as $0$s. 
Figure~\ref{fig:gamma} shows that $\gamma$ should be set as $0.12$ to get the best performance.

\noindent{\bf Hyperparameters of \SOIREDL}

The beam width $\beta$ and the coefficient of regularization $\lambda$ are two important hyperparameters of \SOIREDL. 
We use grid search to choose $\beta$ from $\{$$10$, $50$, $100$, $300$, $500$, $1000$$\}$ and choose $\lambda$ from $\{$$0$, $10^{-3}$, $10^{-2}$, $10^{-1}$, $1$, $10$$\}$.

\begin{figure}[htbp]
  \includegraphics[width=0.43\textwidth]{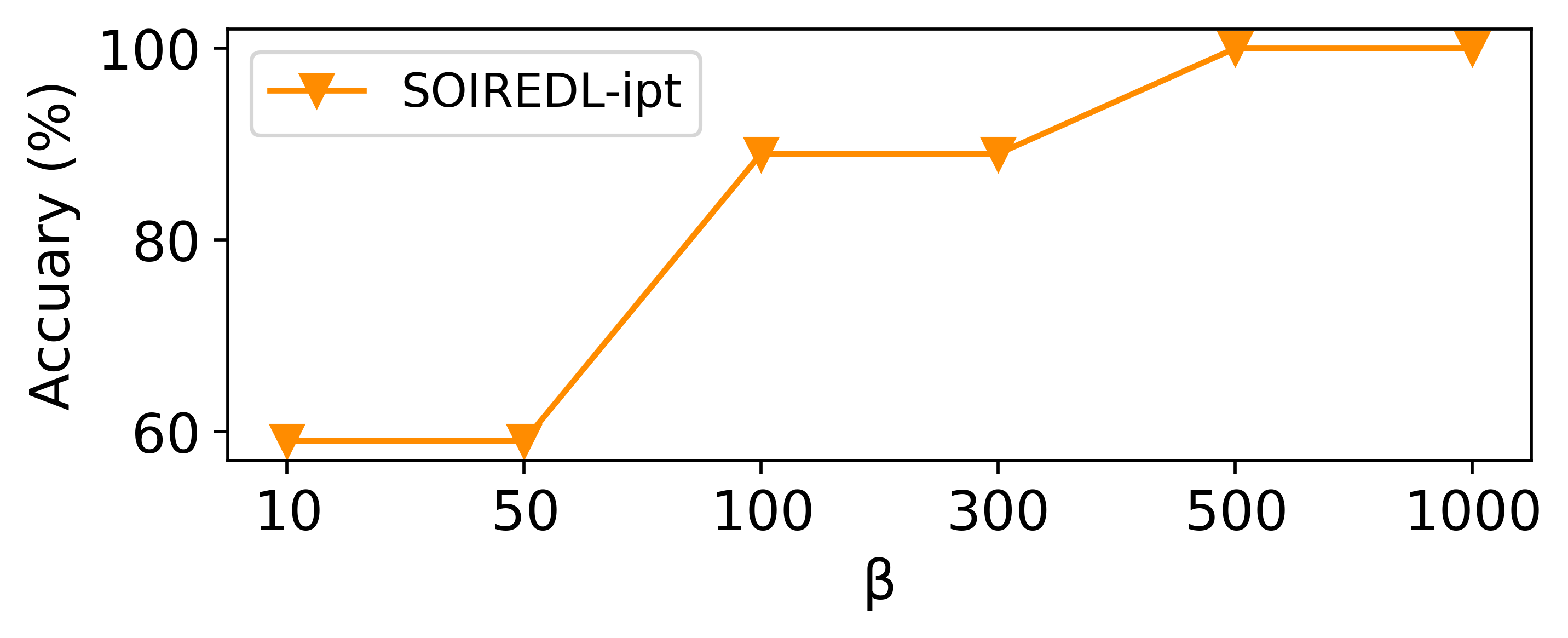}
  \caption{Accuracy(\%) of \SOIREDL on the first dataset with different beam width $\beta$.}
  \label{fig:bw}
\end{figure}

Figure~\ref{fig:bw} shows that \SOIREDL gets the best result when $\beta \geq 500$, so $\beta$ should be set as $500$ to make a balance between accuracy and running time. 

\begin{figure}[htbp]
  \includegraphics[width=0.43\textwidth]{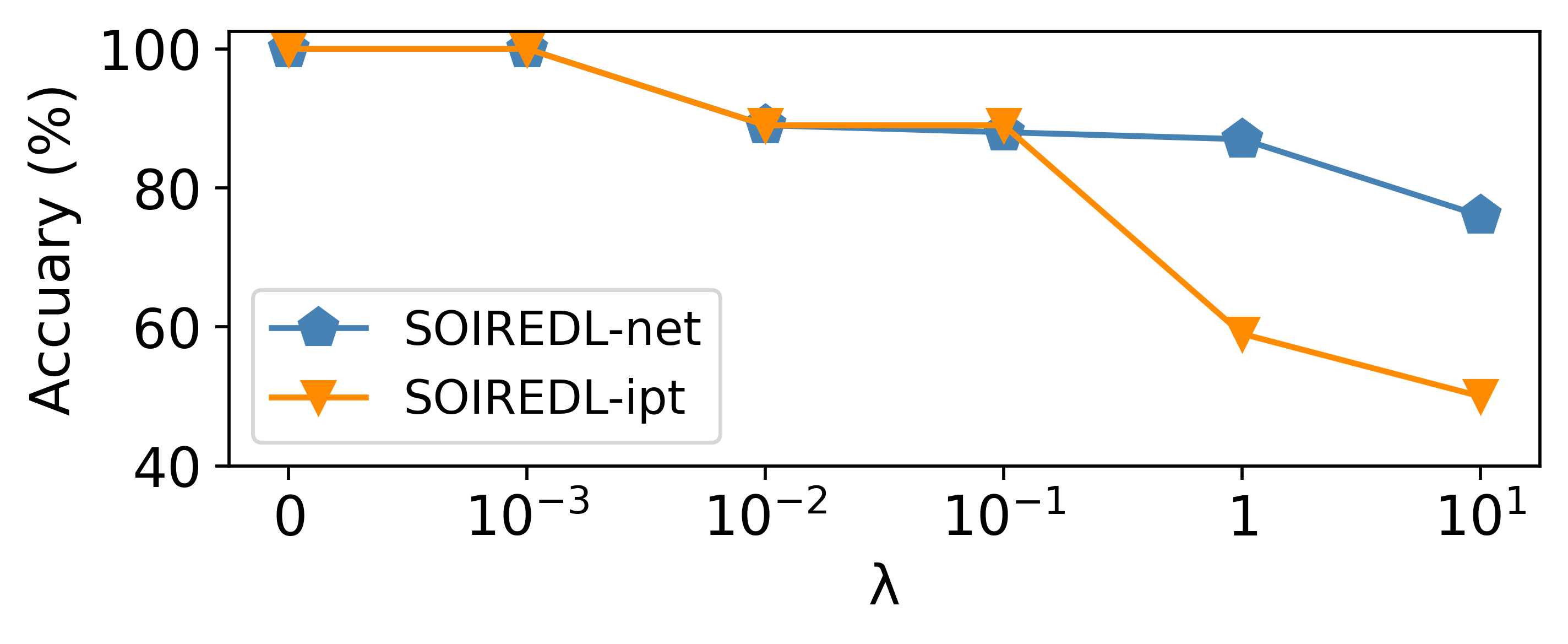}
  \caption{Accuracy(\%) of \SOIREDL on the first dataset with different coefficient of regularization $\lambda$.}
  \label{fig:lambda-acc}
\end{figure}

\begin{figure}[htbp]
  \includegraphics[width=0.43\textwidth]{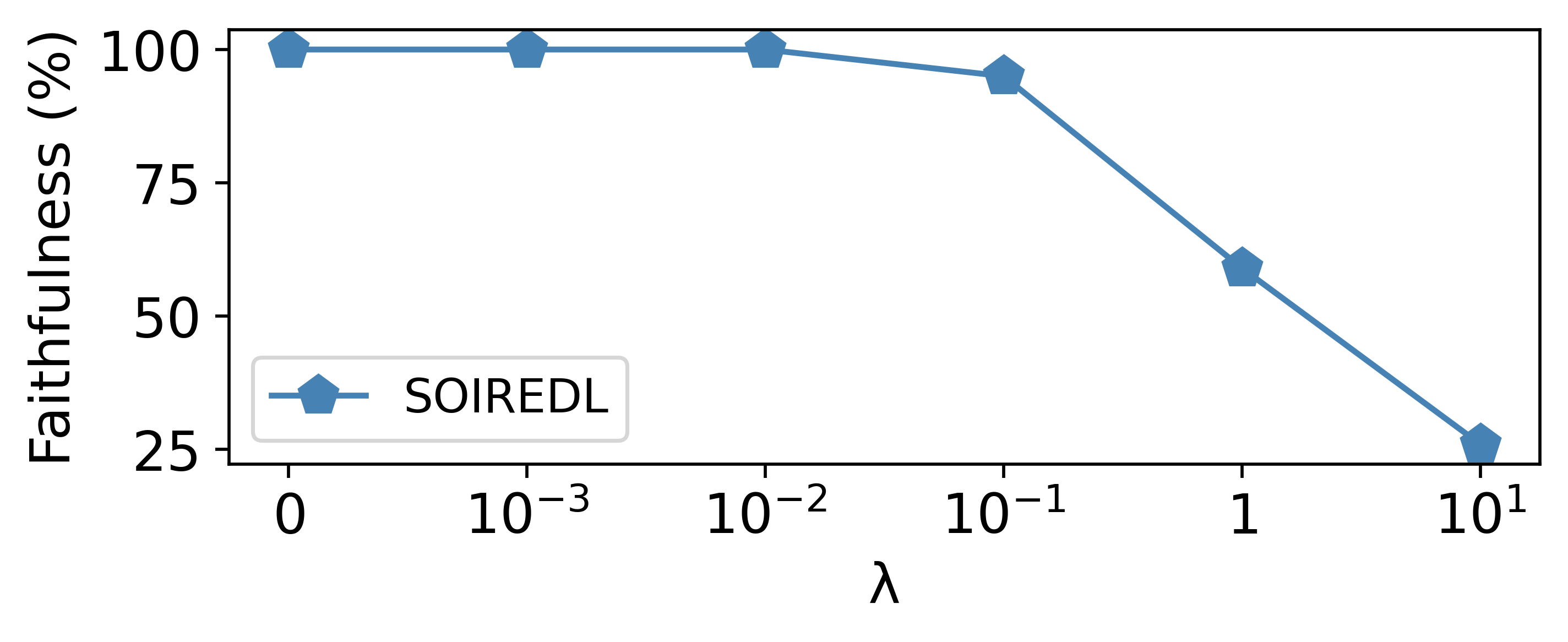}
  \caption{Faithfulness(\%) of \SOIREDL on the first dataset with different coefficient of regularization $\lambda$.}
  \label{fig:lambda-faith}
\end{figure}

Figure~\ref{fig:lambda-acc} and ~\ref{fig:lambda-faith} show that the accuracy and faithfulness decrease when $\lambda>10^{-3}$. 
The larger coefficient of regularization makes the neural network more difficult to train and the results show that \SOIREDL can also gets a well performance without regularization, so we set $\lambda$ as $0$.

\end{document}